\documentclass{article}

\usepackage{neurips_2024}

\usepackage{amssymb}
\usepackage{pifont}
\newcommand{\cmark}{\ding{51}}%
\newcommand{\xmark}{\ding{55}}%

\usepackage[utf8]{inputenc} 
\usepackage[T1]{fontenc}    
\usepackage{hyperref}       
\usepackage{url}            
\usepackage{booktabs}       
\usepackage{amsfonts}       
\usepackage{nicefrac}       
\usepackage{microtype}      
\usepackage{xcolor}         

\usepackage{comment}
\usepackage{amsmath}
\usepackage{graphicx}
\usepackage{subcaption}
\usepackage[ruled,vlined,linesnumbered,noend]{algorithm2e}
\usepackage{amsmath}
\usepackage{amssymb}
\usepackage{mathtools}
\usepackage{amsthm}
\usepackage{pgfplots}   
\usepackage{microtype}
\usepackage{graphicx}
\usepackage{booktabs} 
\usepackage{multirow}

\usepackage{enumitem}
\usepackage{accents} 
\usepackage{caption}
\usepackage{graphicx}

\usepackage[mathscr]{euscript}
\usepackage{cleveref}
\usepackage{wrapfig}

\usepackage{nicematrix}

\title{Federated Dynamical Low-Rank Training with Global Loss Convergence Guarantees\footnotemark[1]}

\author{%
  Steffen Schotth\"ofer,\quad and \quad  M. Paul Laiu \\
  Computer Science and Mathematics Division\\
  Oak Ridge National Laboratory \\
  Oak Ridge, TN 37831 USA \\
  \texttt{\{schotthofers, laiump\}@ornl.gov} \\
}

\usepackage{amssymb}

\newcommand{\norm}[1]{\left\lVert#1\right\rVert}
\newcommand{\abs}[1]{\left|#1\right|}
\newcommand{\set}[1]{\left\lbrace#1\right\rbrace}

\newcommand{\rom}[1]{\uppercase\expandafter{\romannumeral #1\relax}}
\newcommand{\landauO}{\mathcal{O}}

\newcommand{\inner}[1]{\left< #1 \right>}

\newtheorem{theorem}{Theorem}
\newtheorem{lemma}{Lemma}
\newtheorem{corollary}{Corollary}

\newtheorem{assumption}{Assumption}

\newcommand{\augUc}{\widetilde{U}_c}
\newcommand{\augSc}{\widetilde{S}_c}
\newcommand{\augVc}{\widetilde{V}_c}
\newcommand{\augWrc}{\widetilde{W}_{r,c}}
\newcommand{\augU}{{\widetilde{U}}}
\newcommand{\augS}{{\widetilde{S}}}
\newcommand{\hatS}{{\widehat{S}}}

\newcommand{\augV}{{\widetilde{V}}}
\newcommand{\augWr}{{\widetilde{W}_{r}}}
\newcommand{\Wr}{{{W}_{r}}}

\begin{document}

\maketitle
\begin{abstract}In this work, we propose a federated dynamical low-rank training (FeDLRT) scheme to reduce client compute and communication costs - two significant performance bottlenecks in horizontal federated learning. Our method builds upon dynamical low-rank splitting schemes for manifold-constrained optimization to create a global low-rank basis of network weights, which enables client training on a small coefficient matrix. A consistent global low-rank basis allows us to incorporate a variance correction scheme and prove global loss descent and convergence to a stationary point. Dynamic augmentation and truncation of the low-rank bases automatically optimizes computing and communication resource utilization. We demonstrate the efficiency of FeDLRT in an array of computer vision benchmarks and show a reduction of client compute and communication costs by up to an order of magnitude with minimal impacts on global accuracy.
\renewcommand{\thefootnote}{\fnsymbol{footnote}}

\footnotetext[1]{{
This manuscript has been authored by UT-Battelle, LLC under Contract No. DE-AC05-00OR22725 with the U.S. Department of Energy. The United States Government retains and the publisher, by accepting the article for publication, acknowledges that the United States Government retains a non-exclusive, paid-up, irrevocable, world-wide license to publish or reproduce the published form of this manuscript, or allow others to do so, for United States Government purposes. The Department of Energy will provide public access to these results of federally sponsored research in accordance with the DOE Public Access Plan(\url{http://energy.gov/downloads/doe-public-access-plan}).}
\renewcommand{\thefootnote}{\arabic{footnote}}}
\end{abstract}

\section{Introduction}
Federated learning (FL) \cite{kairouz2021advances,li2020federated,shamir2014communication,mcmahan2016communicationefficient} builds a global model on a central \textit{server} from data distributed on multiple devices, i.e., \textit{clients}, by iteratively aggregating local models trained with the computation resource on the clients.
In horizontal FL, where all clients share identical model architecture and data features, computation is often limited by (i) the communication bandwidth between clients and the server and (ii) the restricted compute and memory resources at each client. 
The former could be addressed by deploying various compression techniques, such as sparse randomized sketching~\cite{haddadpour2020fedsketch,kone2017federated}, subsampling~\cite{kone2017federated}, and low-rank approximation~\cite{qiao2021communicationefficient, vogels2020powersgd, 10146425}, or by allowing for partial \cite{mcmahan2016communicationefficient,cho2020client,nishio2019client} or asynchronous \cite{sprague2018asynchronous,chen2020asynchronous} communications.
The latter could be addressed by sparse training \cite{dai2022dispfl,qiu2022zerofl,yang2020federated}, low-rank training~\cite{liu2023differentially,yi2024pfedlora, yao2022fedhm, hyeon-woo2022fedpara, kone2017federated}, and transfer learning \cite{chen2020fedhealth,saha2021federated}.

This work addresses both challenges simultaneously by leveraging dynamical low-rank approximation of the gradient flow with a Galerkin-type operator splitting. It yields a consistent orthogonal low-rank basis across all clients and updates the low-rank factorization without reconstructing the full-weight matrices on either clients or servers. The proposed method yields
1) \textbf{Efficient communication} by only sending and receiving low-rank factors; 2) \textbf{Low client compute and memory footprint}, where the client optimizes only a small coefficient matrix; 3) \textbf{Automatic server-side compression} during training, by augmenting and truncating the weight matrix rank based on the training dynamics. This scheme robustly identifies suitable low-rank manifolds to represent the weight matrices at minimal memory requirements; 4) 
\textbf{Global loss convergence guarantees} to a stationary point of the FL problem, since a globally consistent low-rank basis allows formulation of a variance correction~\cite{NEURIPS2021_7a6bda9a} term to bound each client coefficient drift. 

We demonstrate a significant performance increase in federated scenarios with many clients compared to non-variance corrected methods.

\section{Background and problem statement}

\textbf{Federated optimization}  typically considers \textit{distributed} setups and with \textit{limited communication} and \textit{limited client compute and memory} resources~\cite{mcmahan2016communicationefficient}. 
In this work, we consider a general federated optimization problem, i.e., 
\begin{align}\label{eq_federated_learning}
\textstyle
 \min_{w} \mathcal{L}(w)\,:=\, \frac{1}{C}\sum_{c=1}^C \mathcal{L}_c(w),
\end{align}
where $w$ is a trainable weight, $\mathcal{L}$ is the global loss function associated to a global dataset $X$, and $\mathcal{L}_c$ is the local loss function of client $c$ with local dataset $X_c$ in a federated setup with $C$ clients. 
For notational simplicity, we consider that $X=\cup_{c=1}^C X_c$ and each $X_c$ is of the same size. Therefore, $\mathcal{L}$ is an average of $\mathcal{L}_c$ with uniform weights.
The extension to handle a (non-uniform) weighted average case is straightforward.
\begin{wrapfigure}{r}{0.5\textwidth}
\vspace{-1em}
\centering \includegraphics[width=0.5\textwidth]{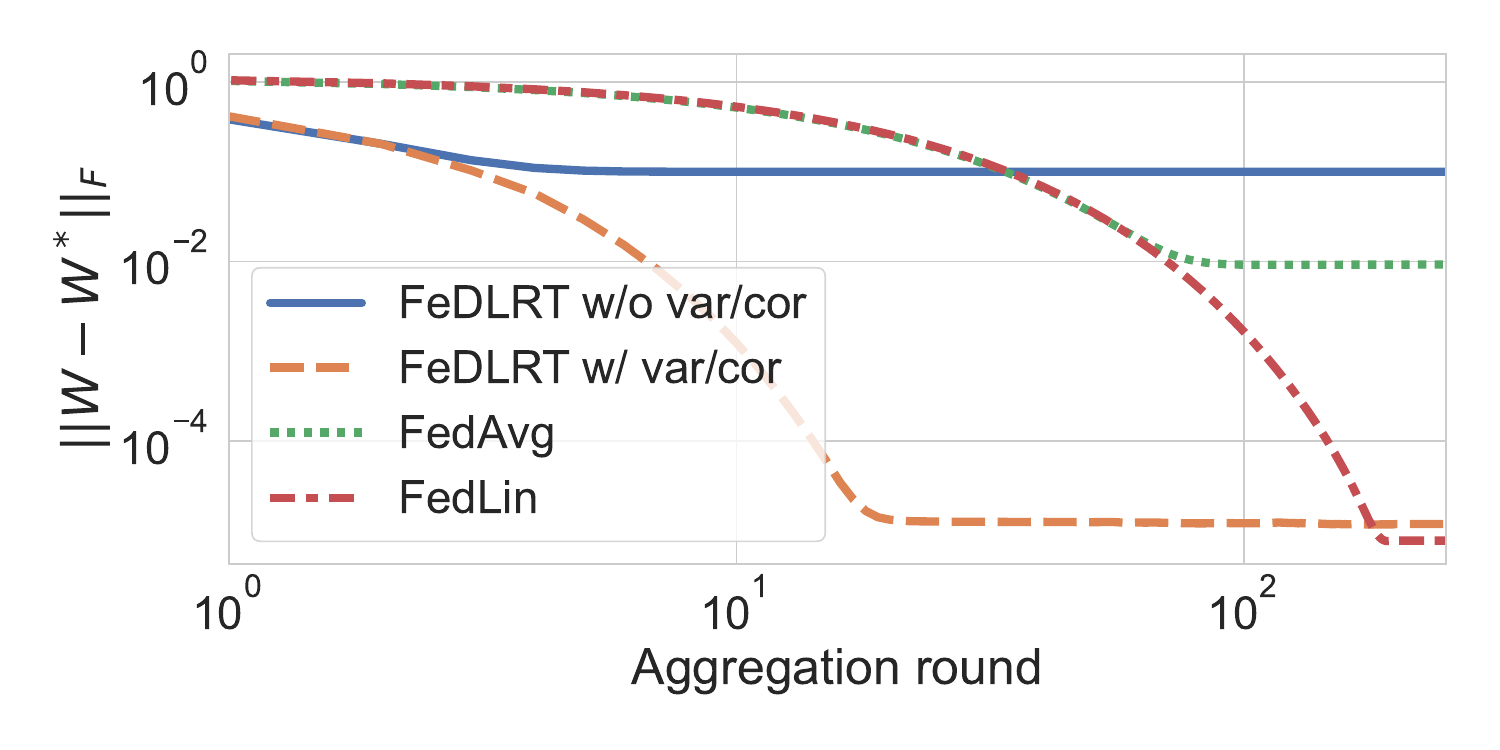}
\caption{Federated, heterogeneous least squares regression problem, see \Cref{subsec:lin_regression}, for $C=4$ clients, $s_*=100$ iterations, learning rate $\lambda=1\rm{e}-3$ and $C$ rank-$1$ local target functions. FL methods without variance correction plateau quickly, whereas FedLin and FeDLRT with variance correction converge to $1\rm{e}-5$. FeDLRT converges faster than FedLin and has lower communication costs.
}
    \label{fig:LSS_multi_obj}
\vspace{0.5em}
\end{wrapfigure}
As the first baseline for federated optimization, we consider FedAvg~\cite{mcmahan2016communicationefficient}, see~\Cref{alg:FedAvg}. Here, each client optimizes its local loss function $\mathcal{L}_c$ for $s_*$ local iterations using gradient descent,
\begin{align}\label{eq_FedAvg_loc_update}
\textstyle
    w_c^{s+1} = w_c^s - \lambda\nabla_w\mathcal{L}(w_c^s),
\end{align}
with learning rate $\lambda$, for $s=0,\dots,s_*-1$. The initial value for the local iteration is the last global weight, i.e., $w_c^{0}=w^t$. After local iterations, the weights are communicated to and aggregated at the server to update the global weight following
\begin{align}\label{eq_FedAvg_aggr}    
\textstyle
    w^{t+1} = \frac{1}{C}\sum_{c=1}^C  w_c^{s_*}.
\end{align}
\textbf{Client-drift effect} is a common challenge in FL, where the iterative client updates \eqref{eq_FedAvg_loc_update} of FedAvg converge to local minima and jeopardize global training performance since the average of the local minimizers may be far away from the global minimizer. 
These effects are particularly pronounced for a large number of local iterations $s_*$, or high discrepancies between local loss functions $\mathcal{L}_c$, as illustrated by \Cref{fig:LSS_multi_obj}. Multiple methods~\cite{shamir2014communication, li2020federated, pathak2020fedsplit, karimireddy2020scaffold, wang2020tackling} have been proposed to mitigate this issue. However, these methods often exhibit a \textit{speed-accuracy conflict}, where learning rates need to be heavily reduced; thus, convergence is slow.

\textbf{Variance correction}\footnote{Variance correction is commonly referred to as ``variance reduction''~\cite{kone2016federated, NEURIPS2021_7a6bda9a}.} introduced in the FedLin method~\cite{NEURIPS2021_7a6bda9a} constructs a variance correction term $V_c =\nabla_w\mathcal{L}_c(w^t) -\frac{1}{C}\sum_{c=1}^C\nabla_w\mathcal{L}_c(w^t)$ and modifies the client update iteration to 
\begin{align}\label{eq_FedLin_loc_update}
    w_c^{s+1} = w_c^s - \lambda\left(\nabla_w\mathcal{L}(w_c^s) -V_c\right) ,\qquad  s=0,\dots,s_*-1.
\end{align}
This technique leads to global convergence to the minimizer of \eqref{eq_federated_learning} with constant learning rates \cite{NEURIPS2021_7a6bda9a} for convex $\mathcal{L}$ and else to convergence to a stationary point, at the cost of an additional communication round for computing the variance correction.

\textbf{Federated neural network training} considers problem \eqref{eq_federated_learning} with the trainable weight $w$ being the set of weight matrices $\set{W_i}_i^L$ of an $L$ layer neural network. In each iteration, the weight updates in \eqref{eq_FedAvg_loc_update} and \eqref{eq_FedLin_loc_update} are applied to all layers simultaneously. Therefore, w.l.o.g., we express the local loss function as $\mathcal{L}_c(W)$, where $W\in\mathbb{R}^{n\times n}$ denotes the weight matrix of an arbitrary layer.

\textbf{Low-rank neural network training:} An array of recent work has provided theoretical and experimental evidence that layer weights of over-parameterized networks tend to be low rank \cite{arora2019implicit, Bah_2022,galanti2022sgd,martin2018implicit} and that removing small singular values may even lead to increased model performance while dramatically reducing model size \cite{sharma2023_laser, Schothoefer_2022} in non-federated scenarios. 
This beneficial feature has spawned a rich landscape of methods to compress neural networks to a low-rank factorization after training with subsequent fine-tuning~\cite{6638949,denton2014exploiting,tjandra2017compressing,lebedev2015speeding}, train the factorized network with fixed rank~\cite{low_rank_conv,wang2021pufferfish,khodak2021initialization}, dynamically adjust the rank during training~\cite{Schothoefer_2022, zangrando2023rankadaptive}, or use low-rank adapters for fine-tuning foundation models~\cite{hu2021lora,dettmers2023qlora,zhao2024galore}. 

\textbf{Why is innovation upon existing low-rank methods needed?}
Since FedAvg~\cite{mcmahan2016communicationefficient}, several low-rank methods~\cite{qiao2021communicationefficient,yi2024pfedlora,liu2023differentially,yao2022fedhm,10146425,hyeon-woo2022fedpara,kone2017federated,reisizadeh2020fedpaq} have been proposed to increase communication and compute efficiency for FL.
These low-rank methods can be categorized into: 1) methods that purely reduce communication cost by communicating only the low-rank factors obtained by performing a full-size SVD (or similar factorization methods) on the full weight matrix after client optimization \cite{qiao2021communicationefficient, vogels2020powersgd, 10146425} and
2) methods that reduce both communication and client compute costs by learning only low-rank factors on clients~\cite{liu2023differentially,yi2024pfedlora, yao2022fedhm, hyeon-woo2022fedpara, kone2017federated}. 

To the best of the authors' knowledge, there is no existing low-rank method that combines 1) efficient communication, 2)
low client compute and memory footprint, 3) automatic server side compression during training, and 4) global loss convergence guarantees using variance correction in the sense of FedLin~\cite{NEURIPS2021_7a6bda9a}, to achieve a globally consistent, robust, and efficient optimization scheme for FL. 

\section{FeDLRT: Federated dynamical low-rank training with variance correction}
In this section, we present the core contribution of this paper, \textit{federated dynamical low-rank training} (FeDLRT), which features a low-rank client optimization step with optional variance correction and an efficient server aggregation process that dynamically determines the optimal weight matrix rank for automatic compression. 

FeDLRT builds on the dynamical low-rank approximation (DLRA) method, which was initially proposed for solving matrix equations \cite{KochLubich07} and recently extended to neural network training~\cite{Schothoefer_2022,zangrando2023rankadaptive, HnatiukPaper}. 

Let $\dot W(t)=-\nabla_W\mathcal{L}(W(t))$ denote the gradient flow for minimizing $\mathcal{L}$.

The DLRA method restricts the trajectory of $W$ to $\mathcal{M}_r$, the manifold of $n\times n$, rank-$r$ matrices, by projecting $\dot W$ onto a local tangent plane of $\mathcal{M}_r$ via an orthogonal projection.
This guarantees a low-rank solution when following the projected dynamics from a low-rank initial guess.

Let the low-rank matrix take the form $W_r =USV^\top\in\mathcal{M}_r$
with $U,V\in\mathbb{R}^{n\times r}$ the orthonormal bases of $\mathcal{M}_r$ and $S\in\mathbb{R}^{r\times r}$ the coefficient matrix. 
The dynamics for each low-rank factor in DRLA are then derived in \cite[Proposition 2.1]{KochLubich07} as
\begin{align}\label{eq:gradient_flow_system}
\textstyle
{
\begin{aligned}
\dot S(t) &= -U^\top(t)\nabla_{W}\mathcal{L}(U(t) S(t)V(t)^\top)V(t), \\
\dot U(t) &= -\left(I - P_{U(t)}\right)\nabla_{W}\mathcal{L}(U(t)S(t)V(t)^\top) V(t)S(t)^{-1}, \\
\dot V(t) &= -\left(I - P_{V(t)}\right)\nabla_{W}\mathcal{L}(U(t)S(t)V(t)^\top) U(t)S(t)^{-\top},
\end{aligned}
}
\end{align}

where $P_{U}=UU^\top$ and $P_{V}=VV^\top$ are the projections onto the column spaces of $U$ and $V$, respectively. 
By using the \textit{basis update \& Galerkin} (BUG) scheme~\cite{ceruti2021rank}, \eqref{eq:gradient_flow_system} can be split into a basis update step for $U$ and $V$ and a coefficient update step for $S$.
This splitting scheme allows for dynamic adjustment of the rank via a basis augmentation before the coefficient update step and a basis truncation after the coefficient update, as shown in \cite{Schothoefer_2022}.

\begin{wrapfigure}{l}{0.4\textwidth}
    \centering
\includegraphics[trim=1.2cm 5cm 22.5cm 1cm, clip, width=0.41\textwidth]{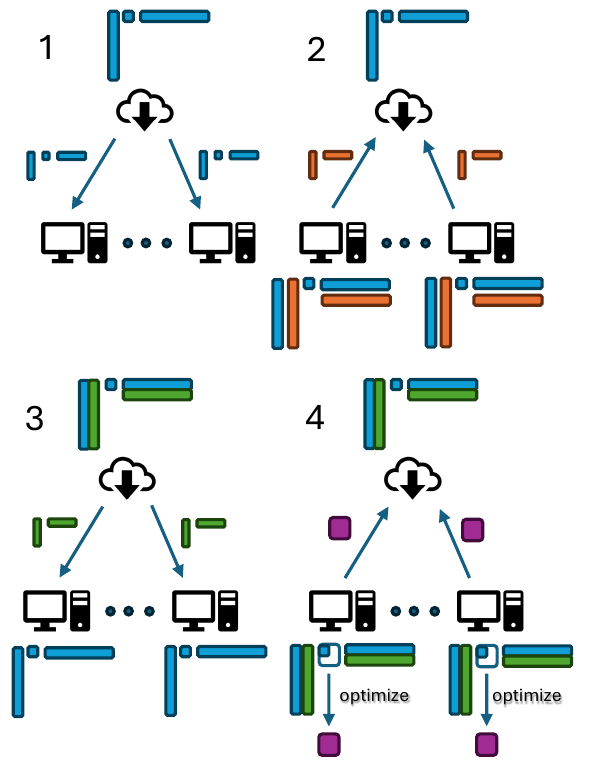}
        \vspace{-2em}
    \caption{Communication of FeDLRT without variance correction. 1) Broadcast current global basis $U,V$ (blue). 2) Aggregate basis gradients $G_{c,U}, G_{c,V}$ (orange). 3) Broadcast global augmented basis $\Bar{U},\Bar{V}$ (green). 4) Aggregate individual client coefficient update $\augSc^{s_*}$(purple). }
 \label{fig:illustration}
  \vspace{-6em}
\end{wrapfigure}

In the context of FL, the BUG splitting scheme is particularly interesting since it allows for learning the low-rank bases and coefficients in separate steps. This gives rise to a globally shared basis for the local client iterations, reducing communication and client compute cost of the proposed FeDLRT scheme, see \Cref{fig:illustration}: First, the factorization is broadcast to the clients (panel 1), and the basis gradients\footnote{and later on the coefficient gradients for variance correction} $U,V$ are aggregated on the server (panel 2). Next, the basis is augmented on the server (panel 3) and broadcast. On the clients, only the augmented coefficient matrix $S$ is updated repeatedly  (panel 4) before aggregation to the server. After aggregation of the local augmented coefficient matrices, redundant basis directions are eliminated to optimize the accuracy-to-compression ratio of the model on the server. 

The strategy yields the following benefits compared to ``full-rank'' FL schemes as FedLin~\cite{NEURIPS2021_7a6bda9a} and low-rank schemes with local compression:\\
\textbf{Low client compute cost:} Server-based basis augmentation and compression enables an automatic compression without a-priori knowledge of the layer rank $r$ and at no cost for the resource-constrained clients. 
The clients only evaluate gradients of low-rank factors and optimize the small matrix $S\in\mathbb{R}^{r\times r}$. 
When the clients are equipped with GPUs, this further implies that all ``GPU unfriendly'' parts of the low-rank scheme, i.e., SVD and QR decomposition for augmenting and compressing the representation, are performed on the server. \\
\textbf{Efficient communication:} Similar to FedLin, FeDLRT requires \textit{in practice} two communication rounds -- one for aggregating and distributing global gradients for basis augmentation and variance correction and one for aggregating locally updated coefficients. However, communication cost for each round is significantly reduced since only low-rank factors are communicated. We refer to \Cref{subsec_com_cost} on communication and compute cost.
Existing federated low-rank schemes effectively generate individual and incompatible representations of $\Wr\in\mathcal{M}_r$ for each client. While the factors can still be efficiently communicated, averaging on the server requires a reconstruction of the full weigh matrix $W^*=\frac{1}{C}\sum_{c=1}^C U_cS_cV_c^\top$, since the local manifolds possibly diverge. Thus, the local rank information is lost and needs to be costly recovered by a full $n\times n$ SVD on the server; see \Cref{alg:naive_FeDLRT} for details. Since the average of low-rank matrices is not necessarily of low rank, these schemes may lose crucial information on the manifold if client solutions drift too far apart from each other. FeDLRT, in contrast, provides the advantage of \textbf{client-wide manifold consistency:} Splitting the low-rank update and sharing bases amongst clients provides a globally consistent manifold basis. This furthermore allows for bounding the coefficient drift, see \Cref{lem_drift_bound_full}, and enables a variance correction for the federated low-rank similar to the FedLin scheme.

\subsection{Description of \Cref{alg:FeDLRT_variance_reduction} - FeDLRT}

In this section, we elaborate on the details in \Cref{alg:FeDLRT_variance_reduction}.
The orthonormal factors $U^t,V^t$ and the coefficient matrix $S^t$ are initialized with rank $r$ and then broadcast to the clients. Note that FeDLRT ensures that, for all $t>1$, $U^t$ and $V^t$ are orthonormal, and $S^t$ is diagonal and full rank.

\textbf{Basis augmentation} of the bases $U^t$ and $V^t$ is performed using concatenation with the corresponding global basis gradients $G_U = \frac{1}{C} \sum_{c=1}^C\nabla_U\mathcal{L}_c(U^tS^tV^{t,\top})$ and $G_V = \frac{1}{C} \sum_{c=1}^C \nabla_V\mathcal{L}_c(U^tS^tV^{t,\top})$, obtained by aggregating the local basis gradients. 
$G_U$ and $G_V$ encapsulate the gradient flow dynamics \eqref{eq:gradient_flow_system} projected onto the original bases, thus yielding an intuitive choice for basis augmentation. Further, this choice is consistent with the basis update step of the augmented BUG splitting scheme, see \Cref{sec:grad_dynamics}, which ensures the robustness of the client optimizer. 
Subsequent orthonormalization, e.g., by a QR decomposition, yields the augmented basis, i.e.,
\begin{align}\label{eq_basis_augmentation}
\textstyle
[U^t \mid \Bar{U}]R = \texttt{qr}([U^t \mid G_U])\in\mathbb{R}^{n\times 2r}, \quad\text{and}\quad [V^t \mid \Bar{V}]R= \texttt{qr}([V^t \mid G_V])\in\mathbb{R}^{n\times 2r}.
\end{align}
We denote the augmented bases by $\augU=[U^t \mid \Bar{U}]$ and $\augV=[V^t \mid \Bar{V}]$.
The orthonormalization is performed on the server, providing compute cost reduction for the client.

\textbf{Basis broadcasting} of $\augU$ and $\augV$ only requires to broadcast the new bases $\Bar{U}$ and $\Bar{V}$, since $U^t$ and $V^t$ are readily available on the clients. Formally, the coefficients $S^t$ are projected onto the augmented basis, i.e., $
    \augS=\augU^\top U^t S^t V^{t,\top} \augV \in\mathbb{R}^{2r\times 2r}$, 
before broadcasting them to the clients. Exploiting the orthonormality of the basis results in further reduction of the communication and compute cost:
\begin{lemma}\label{lem_aug_coeff}
    $\augS=\augU^\top U^t S^t V^{t,\top} \augV$ takes the form  $\augS =
\begin{bmatrix}
S^t & 0 \\
0 & 0
\end{bmatrix}$.
\end{lemma}
See~\Cref{sec:basis_com} for the proof. 
With \Cref{lem_aug_coeff}, only $\Bar{U}$ and $\Bar{V}$ have to be broadcast, and the augmented bases and coefficients $\augU$, $\augV$, and $\augS$ can be assembled on each client as needed. Furthermore, only $S\in\mathbb{R}^{r\times r}$, instead of $\augS\in\mathbb{R}^{2r\times 2r}$, needs to be communicated.

Below, we discuss three options for the client coefficient update step.

\textbf{Client coefficient update} without variance correction is implemented similarly to FedAvg~\eqref{eq_FedAvg_aggr}. On each client $c$, the augmented coefficient matrix $\augS_c$ is trained for $s_*$  iterations%
\footnote{Our analysis focuses on the case where all clients share the same number of local iterations $s_*$. The analysis can be extended to the case where $s_*$ is client dependent, following an approach similar to the one in \cite{NEURIPS2021_7a6bda9a}.} with learning rate $\lambda$,
\begin{align}
     \augS_c^{s+1} = \augS_c^s - \lambda\nabla_{\augS}\mathcal{L}_c(\augU\augS_c^s\augV^\top) ,\quad  s=0,\dots,s_*-1,\quad \text{with}\qquad  \augS_c^{s=0}=\augS.
\end{align}
\textbf{Client coefficient update with variance correction} is required in certain federated scenarios, e.g., the case considered in \Cref{fig:LSS_multi_obj}. Based on FedLin~\cite{NEURIPS2021_7a6bda9a}, we introduce a correction step for the local coefficient update of FeDLRT.
It extends the above local iteration by another communication round, where the gradient of the augmented coefficients $G_{\augS,c}= \nabla_\augS \mathcal{L}_c(\augU\augS\augV^{\top})$ is computed, aggregated to $ G_\augS = \frac{1}{C}\sum_{c=1}^CG_{\augS,c}$ and subsequently broadcast. This yields a correction term  $V_c = G_\augS - G_{\augS,c}$  for each client $c$ and thus the client iterations read
\begin{align}\label{eq_grad_update_var_full}
\textstyle
       \augS_c^{s+1} = \augS_c^s -\lambda\left( \nabla_{\augS}\mathcal{L}_c(\augU\augS_c^s\augV^\top)  +V_c \right),\quad  s=0,\dots,s_*-1,\quad \text{with}\qquad  \augS_c^{s=0}=\augS.
\end{align}
The correction term results in a bound on the coefficient drift and leads to convergence guarantees for FeDLRT, as detailed in \Cref{sec_theory}.

\textbf{Client coefficient update with simplified variance correction}: 
Empirically, we observe that a simplified variance correction, which only considers the correction term of the \textit{non-augmented} coefficients $S^t$, is sufficient, see \Cref{fig:cifar10_resnet18}. 
The simplified variance correction term takes the form
\begin{align}\label{eq_var_cor_simplified}
\textstyle
   V_c = G_\augS - G_{\augS,c} 
   \approx \check{V}_c := \check{G}_\augS - \check{G}_{\augS,c} = 
\begin{bmatrix}
\nabla_S \mathcal{L}(U^tS^tV^{t,\top}) - \nabla_S \mathcal{L}_c(U^tS^tV^{t,\top}) & 0 \\
0 & 0
\end{bmatrix},
\end{align}
which makes lines 10 and 12 in \Cref{alg:FeDLRT_variance_reduction} redundant, since $\check{G}_\augS$ can be aggregated in one step with the basis gradients $G_U$,$G_V$ in line 4 and broadcast with $\Bar{U},\Bar{V}$ in line 6, reducing the communication rounds to two - the same as FedLin. See \Cref{alg:modified_FeDLRT_var_cor} for details.

\textbf{Coefficient averaging} is performed after (any of the above variants of) the client iterations. The server computes the updated global coefficients by averaging the local updates, i.e., $\augS^{*} = \frac{1}{C} \sum_{c=1}^C \augS_c^{s_*}$. 
With the shared augmented bases $\augU$ and $\augV$, this is equivalent to the FedAvg aggregation
\begin{align}
\textstyle
    \augWr^*=  \frac{1}{C} \sum_{c=1}^C\augWr^{s_*} =\frac{1}{C}\sum_{c=1}^C \left(\augU  \augS_c^{s_*} \augV^\top\right) = \augU(\frac{1}{C}\sum_{c=1}^C  \augS_c^{s_*} )\augV^\top =\augU\augS^*\augV^\top.
\end{align}
Since the basis is fixed, the rank $2r$ is preserved in the aggregation, which is
in contrast to other federated low-rank schemes where the aggregated weights could be full rank and, in turn, require a full matrix SVD to determine the new rank \cite{qiao2021communicationefficient,10146425}.

\textbf{Automatic compression via rank truncation} is necessary 1) to identify the optimal rank of the weight matrix and 2) to ensure that $S$ is full rank\footnote{Full rank $S$ is required to show consistency of the basis update step \eqref{eq_basis_augmentation} with the robust operator splitting of \cite{ceruti2021rank, Schothoefer_2022}, see \Cref{sec:grad_dynamics}.}. 
To this end, a truncated SVD of $\augS^{*} \in\mathbb{R}^{2r\times 2r}$ is performed, i.e.
$P_{r_1}, \Sigma_{r_1}, Q_{r_1}^\top = \texttt{svd}(\augS^{*})$, where $P_{r_1},Q_{r_1}\in\mathbb{R}^{2r\times r_1}$ and $\Sigma_{r_1}=\text{diag}(\sigma_{1},\dots,\sigma_{r_1})$ contains the $r_1$ largest singular values of $\augS^{*}$. The new rank $r_1$ can be chosen by a variety of criteria, e.g., a singular value threshold $\norm{[\sigma_{r_1},\dots,\sigma_{2r}]}_2<\vartheta$.
Once a suitable rank is determined, the factorization is updated by the projection of the bases 
$U^{t+1}=\augU P_{r_1}\in\mathbb{R}^{n\times r_1}$, $V^{t+1}=\augV Q_{r_1}\in\mathbb{R}^{n\times r_1}$ and update of the coefficient $S^{t+1}=\Sigma_{r_1}$. Remarkably, \Cref{alg:FeDLRT_variance_reduction} is a federated low-rank learning scheme whose solution is close to a full-rank solution, see \Cref{robust_error_bound}.

\begin{algorithm}[t]
\caption{FeDLRT (See \Cref{alg:aux_funcs} for auxiliary function definitions)}\label{alg:FeDLRT_variance_reduction}
\DontPrintSemicolon
\SetAlgoLined
\SetKwInOut{Input}{Input}
\SetKwComment{Comment}{$\triangleright$\ }{}

\Input{Initial orthonormal bases $U^1,V^1\in\mathbb{R}^{n\times r}$ and full rank $S^1\in\mathbb{R}^{r\times r}$;\;
Client-server setup with clients $c=1,\dots,C$;\;
{\tt var\_cor}: Boolean flag to activate variance correction;\;
$\tau$: singular value threshold for rank truncation.}
\For {$t=1,\dots, T$}{
 {\tt  broadcast}$\left(\set{U^t,V^t,S^t}\right)$ \;
 $G_{U,c}\gets\nabla_U \mathcal{L}_c(U^tS^tV^{t,\top})$;  $G_{V,c}\gets \nabla_V \mathcal{L}_c(U^tS^tV^{t,\top})$\tcc*{On client}
$G_{U},G_{V}\gets$ {\tt  aggregate}$\left(\set{G_{U,c},G_{V,c}}\right)$ \;
 $\Bar{U}\gets${\tt  basis\_augmentation}($U^t$, $G_U$);  $\Bar{V}\gets${\tt  basis\_augmentation}($V^t$, $G_V$) \;
 {\tt  broadcast}$(\{\Bar{U},\Bar{V}\})$ \;
$\widetilde{U}\gets [U^t\mid \Bar{U}]$;   $\widetilde{V} \gets [V^t\mid \Bar{V}]$ \tcc*{Basis assembly on client}
$\widetilde{S}^{s=0}\gets \begin{bmatrix}
S^t & 0 \\
0 & 0
\end{bmatrix} $ \tcc*{Coefficient matrix assembly on client}
\If (){\tt var\_cor} {
 $G_{\augS,c}\gets\nabla_\augS \mathcal{L}_c(\augU\augS\augV^{\top})$\tcc*{Augmented gradient on client}
$G_{\augS}\gets$ {\tt  aggregate}$(\{G_{\augS,c}\})$ \;
 {\tt  broadcast}$(\{G_{\augS}\})$ \;
 {\tt  coefficient\_update\_var\_cor}$(c,\,G_\augS-G_{\augS,c})$\tcc*{On client}
}
\Else{
 {\tt  coefficient\_update}$(c)$\tcc*{On client}
}
$\widetilde{S}^*\gets$  {\tt  aggregate}$(\{ \widetilde{S}_c^{s_*}\})$ \;
$P_{r_1}, \Sigma_{r_1}, Q_{r_1} \gets$ \texttt{svd}$(\widetilde{S}^*)$ with threshold  $\vartheta$ \tcc*{Compression step}
$U^{t+1}\gets   \widetilde{U}P_{r_1}$;
$V^{t+1}\gets   \widetilde{V}Q_{r_1}$;
$S^{t+1}\gets \Sigma_{r_1}$  \tcc*{Basis and coefficient update}
}

\end{algorithm}

\subsection{Analysis of FeDLRT with variance correction}\label{sec_theory}
In this section, we analyze the FeDLRT algorithm under the general assumption that $\mathcal{L}_c$ and $\mathcal{L}$ are $L$-smooth with constant $L$.
Theorems~\ref{theo_loss_descent_full_var_cor} and \ref{theo_glob_convergence} give the convergence results for FeDLRT with full variance correction \eqref{eq_grad_update_var_full} in \Cref{alg:FeDLRT_variance_reduction}. \Cref{theo_loss_descent_red_var_cor} and \Cref{theo_glob_convergence_simple} provide the convergence for FeDLRT with simplified variance correction in \eqref{eq_var_cor_simplified}, as detailed in \Cref{alg:modified_FeDLRT_var_cor}, under additional assumptions given therein.
We note that the analysis does not require convexity of $\mathcal{L}_c$ or $\mathcal{L}$.

\paragraph{FeDLRT convergence with full variance correction.} The variance-corrected client iteration \eqref{eq_grad_update_var_full} leads to the following bound the client coefficient drift.
\begin{theorem}\label{lem_drift_bound_full} 
Given augmented basis and coefficient matrices $\augU$, $\augV$, and $\augS$. 
If the local learning rate $0<\lambda\leq\frac{1}{Ls_*}$ with $s_*\geq 1$ the number of local steps, for all clients $c$,
    \begin{align}
    \textstyle
        \|{\augSc^s-\augSc}\|\leq \exp(1)s_* \lambda\|{\nabla_\augS \mathcal{L}(\augU\augS\augV^\top)}\|, \quad\textup{for}\quad s=1,\dots,s^*-1,
    \end{align}
where $\augSc^s$ is the variance corrected coefficient as given in \eqref{eq_grad_update_var_full}.
\end{theorem}
The critical ingredient for the proof, provided in \Cref{sec:var_reduction_FeDLRT_full}, is the globally shared augmented bases. \Cref{lem_drift_bound_full} bounds the drift of the low-rank representations of the local weight, which gives rise to the following global loss descent guarantee.

\begin{theorem}\label{theo_loss_descent_full_var_cor}
Let $U^t S^t V^{t,\top}$ and $U^{t+1} S^{t+1} V^{t+1,\top}$ be the factorization before and after iteration $t$ of 
\Cref{alg:FeDLRT_variance_reduction} with variance correction and singular value truncation threshold $\vartheta$.  Let the local learning rate be $0<\lambda\leq\frac{1}{12 Ls_*}$, then the global loss descent is bounded by
\begin{align}\label{eq_descent_bound_var_corr_full_trunc}
\begin{aligned}
          \mathcal{L}(U^{t+1} S^{t+1} V^{t+1,\top}) - \mathcal{L}(U^t S^t V^{t,\top}) \leq
          - s_*\lambda(1- 12 s_*\lambda L) \|{\nabla_\augS \mathcal{L}(\augU\augS\augV^\top)}\|^2 + L\vartheta.
\end{aligned}
     \end{align}
     \end{theorem}

The proof is provided in \Cref{sec_app_proof_theo_bound_deterministic_variance_reduction_full}. 
\Cref{theo_loss_descent_full_var_cor} paves the way for the following result on convergence to a global stationary point.
\begin{table}[]
    \centering
        \caption{Comparison of the computational footprint of FeDLRT with FedAvg, FedLin and several low-rank FL methods. The FeDLRT variants are the only low-rank schemes with linearly scaling (in $n$) memory, compute, and communication costs with automatic compression and variance correction.}
 \resizebox{\textwidth}{!}
 {
\begin{NiceTabular}{l | l l l l l c c c }[colortbl-like]
\toprule
 Method & Client compute & Client memory & Server compute & Server memory & Com. Cost & Com. Rounds & var/cor. & rank adaptive\\
\midrule
FedAVG~\cite{mcmahan2016communicationefficient} & $\landauO(s_*bn^2)$ & $\landauO(2n^2)$ & $\landauO(n^2)$ & $\landauO(2n^2)$ & $\landauO(2n^2)$ & 1 & \xmark & \xmark\\
FedLin~\cite{NEURIPS2021_7a6bda9a} & $\landauO(s_*bn^2)$ & $\landauO(2n^2)$ & $\landauO(n^2)$ & $\landauO(2n^2)$ & $\landauO(4n^2)$ & 2 & \cmark & \xmark\\
FeDLRT w/o var/cor & $\landauO(s_*b(4nr+4r^2))$ & $\landauO(4(nr + 2r^2))$ & $\landauO(2nr + (8 + 4n)r^2 + 8r^3  )$ & $\landauO(2nr + 4r^2)$ &  $\landauO(6nr + 6r^2))$ &  2 & \xmark & \cmark\\
FeDLRT simpl. var/cor &  $\landauO(s_*b(4nr+4r^2) + r^2)$ & $\landauO(4(nr + 2r^2))$ & $\landauO(2nr + (8 + 4n)r^2 + 8r^3  )$ & $\landauO(2nr + 4r^2)$ &  $\landauO(6nr + 8r^2)$ &  2 & \cmark & \cmark\\
FeDLRT full var/cor &  $\landauO(s_*b(4nr+4r^2) + 4r^2)$ & $\landauO(4(nr + 2r^2))$ & $\landauO(2nr + (8 + 4n)r^2 + 8r^3  )$ & $\landauO(2nr + 4r^2)$ &  $\landauO(6nr + 10r^2)$ &  3 & \cmark & \cmark\\
FeDLR~\cite{qiao2021communicationefficient} & $\landauO(s_*bn^2 + n^3)$ & $\landauO(2n^2)$  &$\landauO(n^2 + n^3)$ & $\landauO(4nr)$ & $\landauO(4nr)$ & 1 & \xmark & \cmark \\
Riemannian FL~\cite{10146425} & $\landauO(2n^2r + 4n r^2 + 2nr)$ & $\landauO(2n^2)$ & $\landauO(2nr + n^2 r)$ & $\landauO(4nr)$ & $\landauO(4nr)$ & 1 & \xmark & \cmark\\
\bottomrule
\end{NiceTabular}

}
    \label{tab:com_cost}
\end{table}
\begin{theorem}\label{theo_glob_convergence}
      Algorithm ~\ref{alg:FeDLRT_variance_reduction} guarantees that, for learning rate $\lambda\leq\frac{1}{12 Ls_*}$ and final iteration $T$,
    \begin{align}
          \min_{t=1,\dots,T}\norm{\nabla_\augS\mathcal{L}(U^t S^t V^{t,\top})}^2\leq \frac{48 L}{T}\left(\mathcal{L}(U^{1}S^{1}V^{1,\top})-\mathcal{L}(U^{{T+1}}S^{{T+1}}V^{{T+1},\top})\right) + {48 L^2}{\vartheta}.
    \end{align}
\end{theorem}

\begin{figure}
    \centering
       \begin{subfigure}[t]{0.24\textwidth}	
        \centering
        \includegraphics[width=\textwidth]{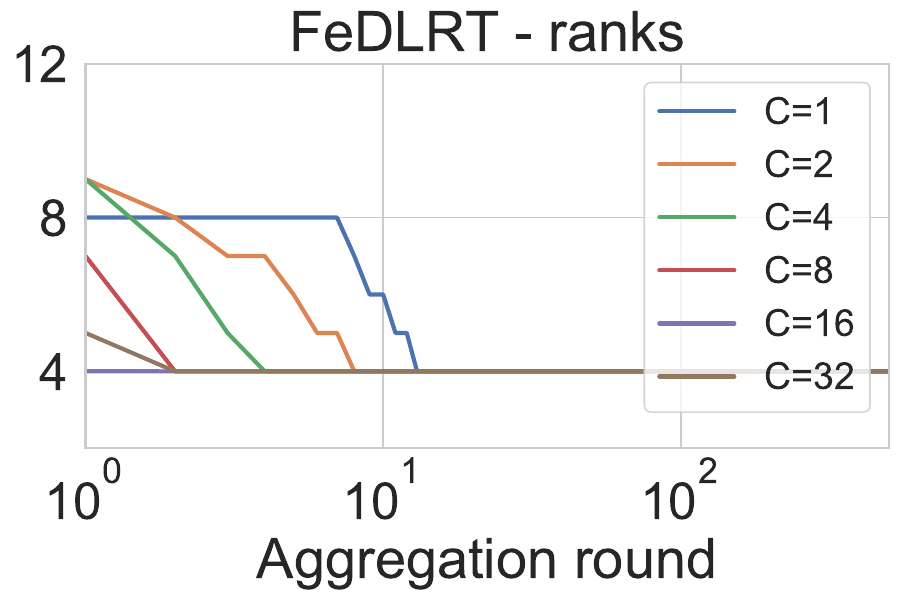}
    \end{subfigure}
    \begin{subfigure}[t]{0.24\textwidth}	
        \centering
        \includegraphics[width=\textwidth]{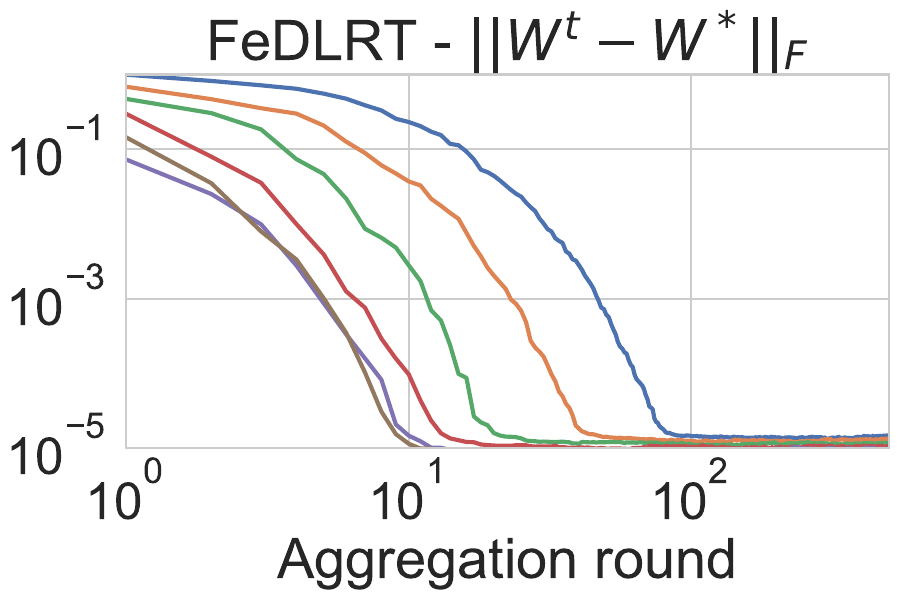}
    \end{subfigure}
\begin{subfigure}[t]{0.24\textwidth}	
        \centering
        \includegraphics[width=\textwidth]{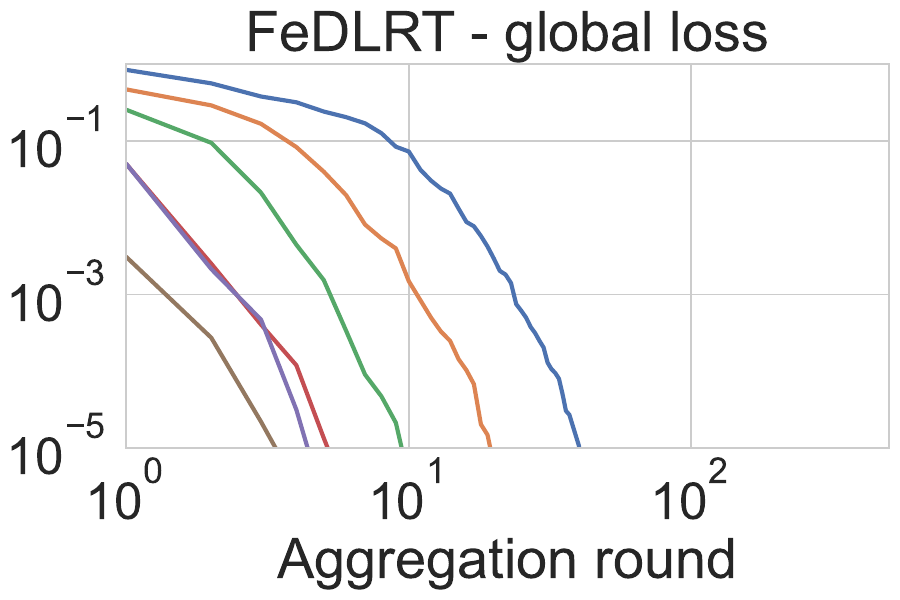}
    \end{subfigure}
    \begin{subfigure}[t]{0.24\textwidth}	
        \centering
        \includegraphics[width=\textwidth]{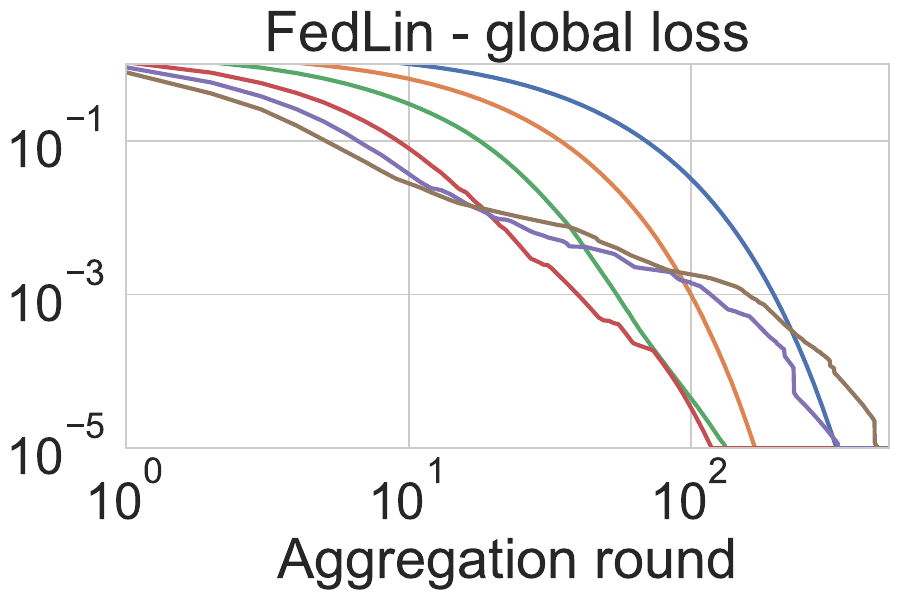}
    \end{subfigure}
    \caption{Comparison between FeDLRT with simplified variance correction and FedLin in the homogeneous linear least squares regression test.  Each line represents the median result of $20$ random initialization with $C$ clients.
    The plots from left to right show the rank evolution, the distance to the global optimizer, the global loss values by FeDLRT, and the global loss values by FedLin.
    The results show that FeDLRT converges faster in this low-rank test case by identifying (and never underestimating) the target rank $r=4$ early in the training.}
    \label{fig:convex}
    \vspace{-1em}
\end{figure}

The proof is given in \Cref{sub_global_convergence}. In particular, this theorem implies convergence 
of \Cref{alg:FeDLRT_variance_reduction} for $T\rightarrow\infty$ up to a $\vartheta$-distance to a global stationary point. This is consistent with the numerical results in \Cref{fig:LSS_multi_obj}, where FedLin converges to the global minimizer (the only stationary point) while FeDLRT with variance correction stops at a point with slightly higher loss value due to a nonzero $\vartheta$. In the case that the FL problem has a low-rank solution, the truncation error bounded by $\vartheta$ vanishes, and convergence to a stationary point is guaranteed, see, e.g., \Cref{fig:convex}.

\begin{wrapfigure}{l}{0.25\textwidth}
    \centering   
     \vspace{-1em}
  \includegraphics[width=0.25\textwidth]{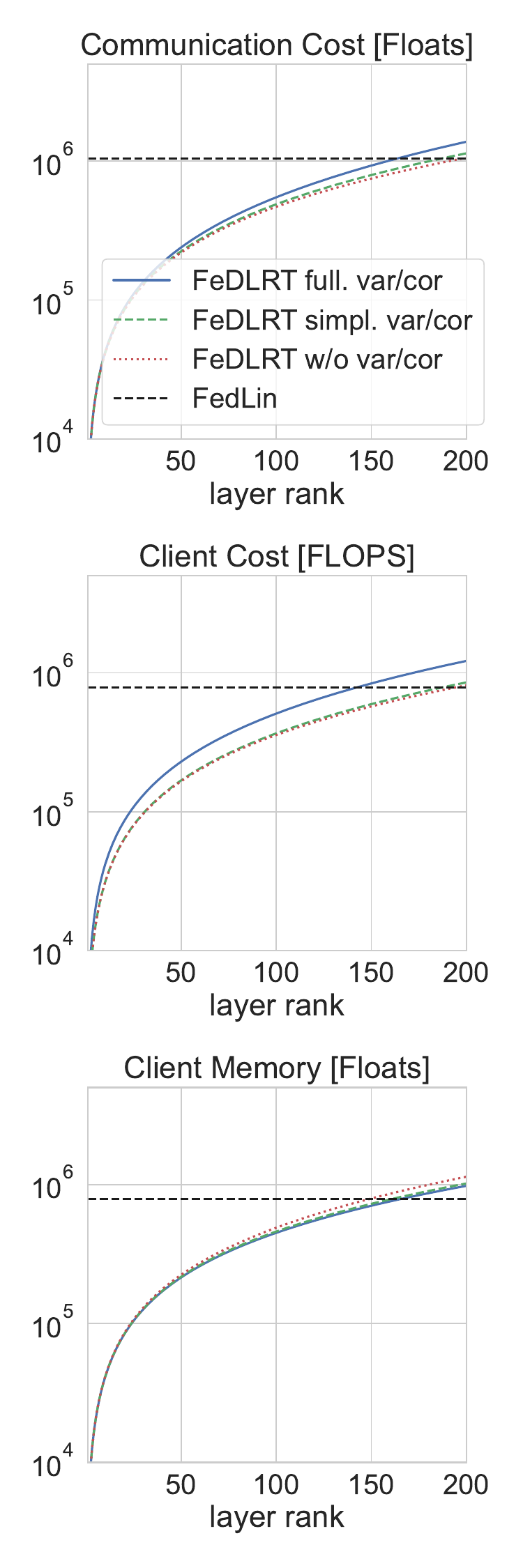}
        \vspace{-2em}
    \caption{Scaling of communication cost (top) compute cost at a single client (middle), and client memory footprint (bottom) for $s_*=1$ client iteration and a single data-point for $W\in\mathbb{R}^{n\times n}$ with $n=512$. The costs drop by orders of magnitude after the amortization point of $r\approx 200$, which is $40\%$ of full rank. The numerical evaluations in \Cref{sec:experiments} show that, in practice, the matrix ranks are typically below the amortization threshold.}
 \label{fig:compute_cost}
  \vspace{-7em}
\end{wrapfigure}
\paragraph{FeDLRT convergence with simplified variance correction.} 
FeDLRT with simplified variance correction is detailed in \Cref{alg:modified_FeDLRT_var_cor} with the variance correction term given in \eqref{eq_var_cor_simplified}, which makes variance correction more communication and computation efficient but comes at a cost of the following additional assumption for convergence analysis. 
\begin{assumption}\label{assump:delta_bound}
There exists $\delta \ll 1$ such that, at each client coefficient update, 
\begin{equation}
\|{\nabla_\augS\mathcal{G}(\augU\augSc^s\augV^\top) }\|
-\|{\nabla_S\mathcal{G}(\augU\augSc^s\augV^\top) }\|
< \delta\|{\nabla_{\augS}\mathcal{L}(\augU\augS\augV^\top)}\|,
\end{equation}
for functions $\mathcal{G} = \mathcal{L}$ and $\mathcal{G} = \mathcal{L}_c$, $c=1,\dots,C$.
\end{assumption}
This assumption can be interpreted as that most of dynamics in the gradient flow are captured in the coefficient update for the original rank-$r$ matrix $S$, and the basis augmentation provides little information. This scenario occurs when FeDLRT identifies the optimal rank, which could happen early for simpler problems as shown in \Cref{fig:convex}, or when FeDLRT approaches a stationary point.

\begin{theorem}\label{theo_loss_descent_red_var_cor}
Under \Cref{assump:delta_bound}, if the local learning rate $0<\lambda\leq\frac{1}{12 Ls_*}$, then \Cref{alg:modified_FeDLRT_var_cor} leads to the global loss descent
\begin{align*}
\mathcal{L}(U^{t+1} S^{t+1} V^{t+1,\top}) - \mathcal{L}(U^t S^t V^{t,\top})\leq 
 -\mathsf{C} \|{\nabla_{\augS}\mathcal{L}(\augWr)}\|^2 + L\vartheta
\end{align*}
with $\mathsf{C}=s_*\lambda(1-\delta^2 - 12s_*\lambda L + \delta^2\,s_*\lambda)$.
\end{theorem}

The proof is provided in \Cref{sec_app_proof_theo_bound_deterministic_simple_variance_reduction}. 
When $\delta$ is small, this bound is slightly weaker than the one in \Cref{theo_loss_descent_full_var_cor}, which leads to the following corollary.
\begin{corollary}\label{theo_glob_convergence_simple}
      Assume that \Cref{assump:delta_bound} holds. 
      \Cref{alg:modified_FeDLRT_var_cor} guarantees that, for the local learning rate $0<\lambda\leq\frac{1}{s_*(12 L +\delta^2)}$, 
    \begin{align*}
          \min_{t=1,\dots,T}\norm{\nabla_\augS\mathcal{L}(U^t S^t V^{t,\top})}^2\leq&\\ \frac{96 L}{T}(\mathcal{L}(U^{1}S^{1}V^{1,\top})- &\mathcal{L}(U^{{T+1}}S^{{T+1}}V^{{T+1},\top})) + {96 L^2}{\vartheta}.
    \end{align*}
\end{corollary}
The proof is analogous to the one for \Cref{theo_glob_convergence}, see \Cref{sub_global_convergence_simple}.

\subsection{Compute and communication cost}\label{subsec_com_cost}

The proposed FeDLRT methods significantly reduce server and client memory footprint, the required communication bandwidth, as well as the client compute cost compared to various baselines, see \Cref{tab:com_cost}. {On the clients}, FeDLRT provides significant memory and compute efficiency since the optimizer only requires the coefficient gradients for the local iterations. To the best of our knowledge, FeDLRT is the only low-rank method with adaptive compression  incorporating variance correction, whose server compute cost scales linearly with the layer dimension since the SVD for rank truncation only needs to be computed on the augmented coefficient matrix of size $2r\times 2r$. We remark that the complete federated learning process is performed on the low-rank factors, and the full matrix $\Wr$ is never required, as, e.g., in \cite{qiao2021communicationefficient,10146425}. FeDLRT significantly reduces the communication cost between server and clients of a compressed layer and the compute cost on the client and server side, compared to FedLin, see \Cref{fig:compute_cost}, see \Cref{alg:FeDLRT_variance_reduction}. The client cost of the simplified variance corrected FeDLRT, see \Cref{alg:modified_FeDLRT_var_cor} is slightly reduced compared to the full variance corrected since only non-augmented coefficient gradients have to be computed and communicated. However, the asymptotic behavior is the same. Its main advantage is the reduced number of communication rounds.

\section{Numerical evaluation}\label{sec:experiments}
\subsection{Distributed linear least squares regression}\label{subsec:lin_regression}
\paragraph{Homogeneous test.}
We first consider a (convex) FL problem \eqref{eq_federated_learning} for linear least squares regression with local loss $\mathcal{L}_c(W)= \frac{1}{2\abs{X_c}}\sum_{(x,y)\in X_c}\norm{p(x)^\top W p(y) - f(x,y)}_2^2$,
where $W\in\mathbb{R}^{n\times n}$ and $p:[-1,1]\rightarrow \mathbb{R}^n$ is the Legendre polynomial basis of degree $n-1$. 
The target function $f$ is manufactured as $f(x,y)= p(x)^\top W_r p(y)$, where $\text{rank}(W_r)=r$. 
We consider problems with $n=20$, $r=4$, and randomly generated $W_r$, with $10,000$ data points uniformly sampled on $[-1,1]^2$ and uniformly distributed among clients.
We compare FeDLRT with variance correction and FedLin with $s_*=20$ local iterations and $\lambda=1e-3$ learning rate on $C=1,2,4,8,16,32$ clients. This setting satisfies the step-size restriction given in \Cref{theo_loss_descent_full_var_cor}. 
In FeDLRT, the singular value truncation threshold $\vartheta=\tau||\augS^*||$ with $\tau=0.1$ was used.

Figure~\ref{fig:convex} reports the dynamically updated ranks, errors, and loss values with respect to the aggregation rounds. The reported data are the medians over $20$ randomly generated initial weights%
\footnote{We chose to display the median trajectory to point out its convergence and monotonicity. The test case also converges in the mean.} 
The results indicate that FeDLRT is able to identify the correct rank within a few aggregation rounds and, furthermore, never underestimates it -- which would have increased the loss value significantly. 
FeDLRT converges to the minimizer $W^*=W_r$ up to a $1e-5$ error and converges faster with more clients. On this problem, FeDLRT shows up to 10x faster convergence than FedLin. 
We attribute this behavior to the fact that, by identifying a suitable low-rank manifold early in the training, FeDLRT significantly reduces the degrees of freedom in the FL problem.
\paragraph{Heterogeneous test.}
Inspired by~\cite{NEURIPS2021_7a6bda9a}, we consider a variation of the linear least squares regression with $\mathcal{L}_c(W) = \frac{1}{2\abs{X}}\sum_{(x,y)\in X} \norm{p(x)^\top W p(y) - f_c(x,y)}^2$, where the target function $f_c$ is different for each client, and the $10,000$ training data points are available to all clients. We choose problem size $n=10$ with $C=4$ clients and use learning rate $\lambda=1\rm{e}-3$ with $s_*=100$ local epochs. As seen in \Cref{fig:LSS_multi_obj}, FeDLRT with variance correction converges (to single precision accuracy) to the minimizer $W^*$ of \eqref{eq_federated_learning} much faster than FedLin, whereas FeDLRT without correction quickly plateaus, similar to FedAvg.

\begin{figure}
    \centering
  \includegraphics[width=\textwidth]{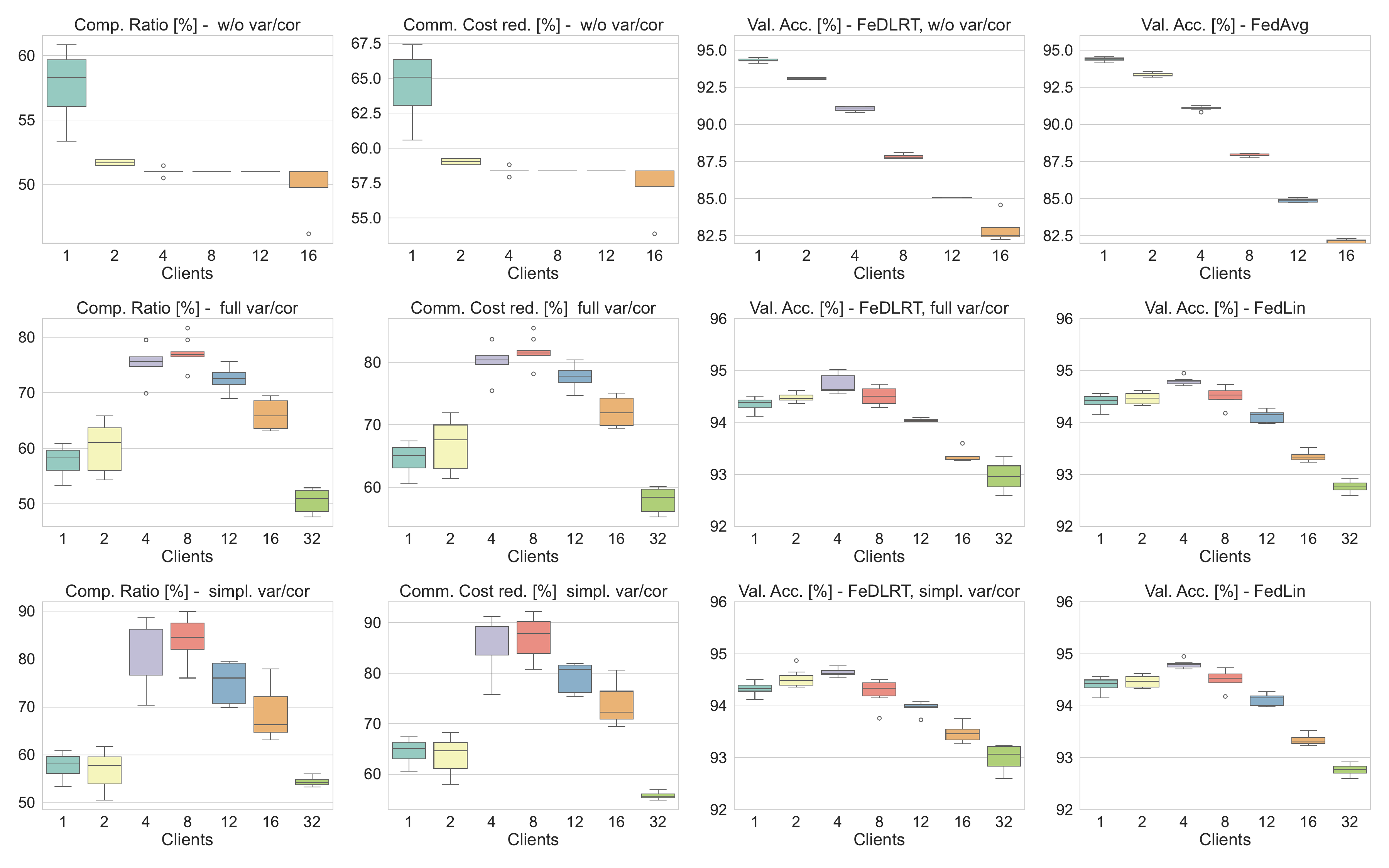}
    \caption{Comparisons for training ResNet18 on CIFAR10 benchmark. Top row compares FeDLRT without variance correction to FedAvg, middle and bottom rows compare FeDLRT with full and simplified variance correction to FedLin, respectively.
    In each row, the left two panels show the model compression ratio and the communication cost reduction from FeDLRT, and the right two panels show the validation accuracy for FeDLRT and the full-rank counterparts. 
    In each plot, the results are reported for $C=1,\dots,16$ or $32$ clients with $240/C$ local iterations.    
    FeDLRT matches the accuracy of FedAvg and FedLin well, while substantially reducing the server and client memory and communication costs. 
    Variance correction leads to an up to $12\%$ increase in validation accuracy for large $C$, mitigating the client drift problem. The simplified variance correction (bottom row) gives comparable results to full version (middle row) at a lower communication and computation cost.}
 \label{fig:cifar10_resnet18}
 \vspace{-1.8em}
\end{figure}
\subsection{ResNet18 on CIFAR10}\vspace{-0.8em}
We demonstrate the performance of FeDLRT for training the exemplary ResNet18 model on CIFAR10, where we apply FeDLRT to train its fully connected head. The truncation tolerance is set to $\vartheta=\tau||\augS^*||$ with $\tau=0.01$.
 The test case setup is summarized in \Cref{tab:object_detection_setup}. The training data is equally partitioned across clients; see \Cref{sec_data} for the data-preprocessing details. 
A local iteration of \Cref{alg:FeDLRT_variance_reduction} at client $c$ describes one mini-batch update on the client training data set $X_c$ for a given batch size, $s_*$ is the maximum number of local iterations, and $T$ denotes the number of aggregation rounds. 
We display the statistics for $10$ random initializations;
 each warm-started with $5$ iterations with one client.
We set $s_*=240/C$ so that in each training run, the global network iterates through the same amount of data. This setup favors low client counts, and, as expected, the validation accuracy drops as $C$ grows for FedAvg and FeDLRT without variance correction, see \Cref{fig:cifar10_resnet18} (upper row). We note that FeDLRT ties or outperforms FedAvg in terms of accuracy. 
Using full variance correction (second row) increases the validation accuracy of FeDLRT by up to $12\%$ in this test case, matching the accuracy of FedLin and enabling FL with $93\%$ accuracy for $32$ clients. For $C=8$ clients, the communication cost saving of the compressed layers is up to $90\%$. 
The computationally more efficient simplified variance correction, using \Cref{alg:modified_FeDLRT_var_cor}, (third row), yields similar validation accuracy, notably at higher compression ratio and communication cost reduction. 
Similar results are obtained for AlexNet, VGG16 on CIFAR10, and ViT on CIFAR100 , see \Cref{sec:add_numerical_results}, where we observe that FeDLRT closely matches the full-rank accuracy of FedLin.

\textbf{In conclusion}, 
 we have presented with FeDLRT an efficient low-rank FL scheme with convergence guarantees and automatic server side compression, and demonstrated its capabilities in several test cases. We remark that the underlying assumption for this work is that the target model can be expressed sufficiently well via a low-rank representation.
While this work is baseline algorithmic research, we believe that it leads to positive societal impacts by providing an energy efficient FL algorithm.

\newpage
\bibliographystyle{abbrv}
\bibliography{lit.bib}

\newpage
\appendix

\section{Additional algorithms}
In the following, we list a set of algorithms that are used in the paper as a contribution or as a baseline method. In particular, \Cref{alg:aux_funcs} contains auxiliary function definitions for \Cref{alg:FeDLRT_variance_reduction} and \Cref{alg:modified_FeDLRT_var_cor}. 
\Cref{alg:FedAvg} is the standard FedAvg method as presented in \cite{mcmahan2016communicationefficient}. \Cref{alg:FedLin} is the FedLin Algorithm \cite{NEURIPS2021_7a6bda9a}, i.e. the extension of \Cref{alg:FedLin} with variance correction.
\Cref{alg:modified_FeDLRT_var_cor} represents the FeDLRT method with simplified variance correction, as analyzed in \Cref{theo_loss_descent_red_var_cor} and \Cref{theo_glob_convergence_simple} with the additional \Cref{assump:delta_bound}.

\begin{algorithm}[ht]
\caption{Auxiliary functions}\label{alg:aux_funcs}

\DontPrintSemicolon
\SetAlgoLined

\vspace{.2em}
\SetKwProg{Def}{def}{:}{}
\Def{   {\tt  broadcast}($\set{M_i}_i$: list of matrices)}{
Send $M_i$ from server to all clients $\forall i$
}
\vspace{.2em}
\SetKwProg{Def}{def}{:}{}
\Def{   {\tt  aggregate}($\set{M_{c,i}}_i$: list of matrices)}{
Send $M_{c,i}$ from client to server  $\forall c,i$\\
$M_i\gets \frac{1}{C}\sum_{c=1}^C M_c\quad \forall i$ \\
return $\set{M_i}_i$;
}
\vspace{.2em}
\SetKwProg{Def}{def}{:}{}
\Def{   {\tt  coefficient\_update\_var\_cor}($c$: client, $V_c$: correction term)}{
\For(\tcc*[f]{On client}){$s=0,\dots, s_*-1$}{
$\widetilde{S}_c^{s+1}\gets \widetilde{S}_c^{s} - \lambda \left(\nabla_\augS\mathcal{L}_c(\widetilde{U}_c\widetilde{S}_c^{s}\widetilde{V}_c^\top)+
V_c\right)$\\
}
}
\vspace{.2em}
\SetKwProg{Def}{def}{:}{}
\Def{   {\tt  coefficient\_update}($c$: client)}{
\For(\tcc*[f]{On client}){$s=0,\dots, s_*-1$}{
$\widetilde{S}_c^{s+1}\gets \widetilde{S}_c^{s} - \lambda \nabla_\augS\mathcal{L}_c(\widetilde{U}_c\widetilde{S}_c^{s}\widetilde{V}_c^\top)$\\
}
}
\vspace{.2em}
\SetKwProg{Def}{def}{:}{}
\Def{   {\tt  basis\_augmentation}($B$: old basis, $G_B$: basis dynamics)}{
$[B \mid \Bar{B}] \gets  \texttt{qr}([B \mid G_B])$   \tcc*{On server}
return  $\Bar{B}$
}
\end{algorithm}

\begin{algorithm}[ht]
\DontPrintSemicolon
\SetAlgoLined
\SetKwInOut{Input}{Input}
\SetKwComment{Comment}{$\triangleright$\ }{}

\Input{Initial values for weight matrix $W$\;
Client-server setup with clients $c=1,\dots,C$.
}
\For {$t=1,\dots,T$}{ 
 {\tt  broadcast}$\left(\set{W^t}\right)$ \;
 $W_c^{s=0}\gets W^t$ \;
\For {$s=0,\dots, s_*-1$}{
$W_c^{s+1}\gets W_c^{s} - \lambda \nabla_W\mathcal{L}_c(W_c^s) $ \tcc*{Gradient descent on client}
}
$W^{t+1}\gets$ {\tt  aggregate}$\left(\set{W_c^{s_*}}\right)$ \tcc*{Aggregation on server}
}

\caption{FedAvg~\cite{mcmahan2016communicationefficient}. (See \Cref{alg:aux_funcs} for auxiliary function definitions)}\label{alg:FedAvg}
\end{algorithm}

\begin{algorithm}[ht]
\DontPrintSemicolon
\SetAlgoLined
\SetKwInOut{Input}{Input}
\SetKwComment{Comment}{$\triangleright$\ }{}

\Input{Initial values for weight matrix $W$\;
Client-server setup with clients $c=1,\dots,C$.
}
\For {$t=1,\dots,T$}{ 
 {\tt  broadcast}$\left(\set{W^t}\right)$ \;
 $G_{W,c}\gets \nabla_W\mathcal{L}_c(W^t)$ \tcc*{Gradient computation on client}
$G_{W}\gets$ {\tt  aggregate}$\left(\set{G_{W,c}}\right)$ \tcc*{Aggregation on server}
 {\tt  broadcast}$\left(\set{G_{W}}\right)$ \;
 $W_c^{s=0}\gets W^t$ \;
 $V_c\gets G_{W} - G_{W,c}$\tcc*{Correction term computation on client} 
\For {$s=0,\dots, s_*-1$}{
$W_c^{s+1}\gets W_c^{s} - \lambda \nabla_W\mathcal{L}_c(W_c^s) +V_c $ \tcc*{Corrected iteration on client}
}
$W^{t+1}\gets$ {\tt  aggregate}$\left(\set{W_c^{s_*}}\right)$ \tcc*{Aggregation on server}
}

\caption{FedLin~\cite{NEURIPS2021_7a6bda9a}. (See \Cref{alg:aux_funcs} for auxiliary function definitions)}\label{alg:FedLin}
\end{algorithm}

\begin{algorithm}[ht]
\DontPrintSemicolon
\SetAlgoLined
\SetKwInOut{Input}{Input}
\SetKwComment{Comment}{$\triangleright$\ }{}

\Input{Initial orthonormal bases $U^1,V^1\in\mathbb{R}^{n\times r}$ and full rank $S^1\in\mathbb{R}^{r\times r}$;\;
Client-server setup with clients $c=1,\dots,C$;\;
$\tau$: singular value threshold for rank truncation.}
\For {$t=1,\dots, T$}{
 {\tt  broadcast}$\left(\set{U^t,V^t,S^t}\right)$ \;
 $G_{U,c}\gets\nabla_U \mathcal{L}_c(U^tS^tV^{t,\top})$\tcc*{On client} 
 $G_{V,c}\gets \nabla_V \mathcal{L}_c(U^tS^tV^{t,\top})$\tcc*{On client}
 $G_{S,c}\gets \nabla_S \mathcal{L}_c(U^tS^tV^{t,\top})$\tcc*{On client}
$G_{U},G_{V}, G_{S}\gets$ {\tt  aggregate}$\left(\set{G_{U,c},G_{V,c}, G_{S,c}}\right)$ \;
 $\Bar{U}\gets${\tt  basis\_augmentation}($U^t$, $G_U$),  $\Bar{V}\gets${\tt  basis\_augmentation}($V^t$, $G_V$) \;
 {\tt  broadcast}$\left(\set{\Bar{U},\Bar{V},G_{S}}\right)$ \;
$\widetilde{U}\gets [U^t\mid \Bar{U}]$,   $\widetilde{V} \gets [V^t\mid \Bar{V}]$ \tcc*{Basis assembly on client}
$\widetilde{S}^{s=0}\gets \begin{bmatrix}
S^t & 0 \\
0 & 0
\end{bmatrix} $ \tcc*{Coefficient matrix assembly on client}
$\check{G}_{\augS,c}\gets \begin{bmatrix}
G_{S,c} & 0 \\
0 & 0
\end{bmatrix} $ \tcc*{Client coeff. gradient approximation on client}
$\check{G}_{\augS}\gets \begin{bmatrix}
G_{S} & 0 \\
0 & 0
\end{bmatrix} $ \tcc*{Global coeff. gradient approximation on client}
 {\tt  coefficient\_update\_var\_cor}$\left(c,\,\check{G}_\augS-\check{G}_{\augS,c}\right)$\tcc*{On client}
$\widetilde{S}^*\gets$  {\tt  aggregate}$\left(\set{ \widetilde{S}_c^{s_*}}\right)$ \;
$P_{r_1}, \Sigma_{r_1}, Q_{r_1} \gets$ \texttt{svd}$(\widetilde{S}^*)$ with threshold  $\vartheta$ \tcc*{Compression step}
$U^{t+1}\gets   \widetilde{U}P_{r_1}$, and
$V^{t+1}\gets   \widetilde{V}Q_{r_1}$ \tcc*{Basis projection}
$S^{t+1}\gets \Sigma_{r_1}$ 
}
\caption{FeDLRT with simplified variance correction. (See \Cref{alg:aux_funcs} for auxiliary function definitions)}\label{alg:modified_FeDLRT_var_cor}
\end{algorithm}

\begin{algorithm}[ht]
\DontPrintSemicolon
\SetAlgoLined
\SetKwInOut{Input}{Input}
\SetKwComment{Comment}{$\triangleright$\ }{}

\Input{Initial orthonormal bases $U^1,V^1\in\mathbb{R}^{n\times r}$ and full rank $S^1\in\mathbb{R}^{r\times r}$;\;
Client-server setup with clients $c=1,\dots,C$;\;
$\tau$: singular value threshold for rank truncation.}
\For {$t=1,\dots, T$}{
 {\tt  broadcast}$\left(\set{U^t,V^t,S^t}\right)$ \;
 $U_c^{s=0},V_c^{s=0},S_c^{s=0}\gets U^t,V^t,S^t$ \;
 \For(\tcc*[f]{On client}){$s=0,\dots, s_*-1$}{

 $G_{U,c}\gets\nabla_U \mathcal{L}_c(U_c^sS_c^sV_c^{s,\top})$\;
 $G_{V,c}\gets \nabla_V \mathcal{L}_c(U_c^sS_c^sV_c^{s,\top})$\;
 $\augUc,\_ \gets \texttt{qr}([U_c^s \mid G_{U,c}])$\;
$\augVc,\_ \gets \texttt{qr}([V_c^s \mid G_{V,c}])$ \;
$\augSc=\augUc^\top U_c^s S_c^s V_c^{s,\top} \augVc$\;
$\augSc^{*}\gets \widetilde{S}_c - \lambda \nabla_\augS\mathcal{L}_c(\widetilde{U}_c\widetilde{S}_c\widetilde{V}_c^\top)$\;

$\widetilde{S}^*\gets$  {\tt  aggregate}$\left(\set{ \widetilde{S}_c^{*}}\right)$ \;
$P_{r_1}, \Sigma_{r_1}, Q_{r_1} \gets$ \texttt{svd}$(\widetilde{S}^*)$ with threshold  $\vartheta$ \tcc*{Compression step}
}
$U^{t+1}\gets   \widetilde{U}P_{r_1}$, and
$V^{t+1}\gets   \widetilde{V}Q_{r_1}$ \tcc*{Basis projection}
$S^{t+1}\gets \Sigma_{r_1}$ 
}
\caption{Naive implementation of FeDLRT. (See \Cref{alg:aux_funcs} for auxiliary function definitions)}\label{alg:naive_FeDLRT}
\end{algorithm}

\section{Additional numerical evaluation}\label{sec:add_numerical_results}

\subsection{Compute resources}\label{sec_compute_resources}
The convex test cases are computed on a single Nvidia GTX1080ti GPU. The computer vision benchmarks use a set of Nvidia Tesla V100-SXM2-16GB and Tesla P100-PCIE-16GB. For prototyping, a Nvidia RTX 4090 is used.

\subsection{Data augmentation}\label{sec_data}
We use standard data augmentation techniques for the proposed test cases. That is, for CIFAR10, we augment the training data set by a random horizontal flip of the image, followed by a normalization using mean $[0.4914, 0.4822, 0.4465]$ and std. dev. $[0.2470, 0.2435, 0.2616]$. The test data set is only normalized. 
The same augmentation is performed for CIFAR100, where with mean $[0.5071, 0.4867, 0.4408]$ and std. dev. $[0.2673, 0.2564, 0.2762]$.

\subsection{Additional computer vision results}\label{sec_a}

\begin{figure}
    \centering
\begin{subfigure}[t]{\textwidth}	
  \includegraphics[width=\textwidth]{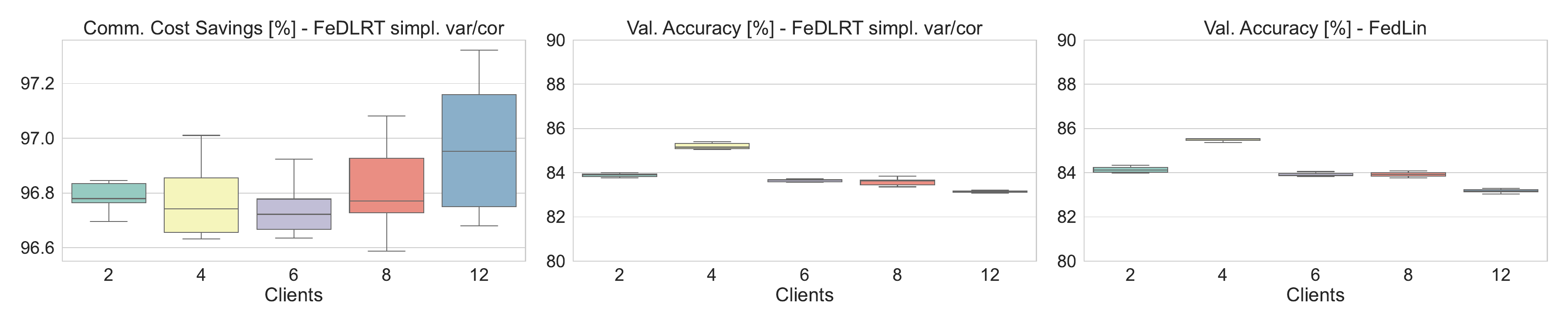}
    \end{subfigure}
    \caption{
  AlexNet CIFAR10 benchmark with fixed number of local iterations. (Left Panel) shows the savings in communication cost of simplified variance corrected FeDLRT vs FedLin. (Mid and right panel) compares the validation accuracy of FeDLRT and FedLin, where we see that FeDLRT behaves similarly to FedLin and achieves accuracy levels near the non-federated baseline value of $85.6\%$.   }
 \label{fig:alex_cifar10}
\end{figure}

\textbf{AlexNet on CIFAR10:} We train AlexNet on CIFAR10, where the fully connected head of the network is replaced by a low-rank counterpart. A federated neural network setup with $C$ clients trains on $CTs_*$ random batches of the dataset, that is the number of seen training data batches scales with the client count. 
\Cref{fig:alex_cifar10} displays the validation accuracy of FeDLRT with variance correction compared to FedLin, where one can see that the performance of FeDLRT mirrors the performance of FedLin with more degrees of freedom. The measured validation accuracy peaks at $C=4$ clients in both cases, where the higher number of seen training data-points offsets the negative effects of more clients on the validation performance. All reported runs are within close distance of the non-federated, full-rank baseline accuracy of $85.6\%$. Communication cost savings of the fully connected layers amount between $96\%$ and $97\%$ \footnote{For clarity of exposition we consider only the fully connected layers. Taking into account the non low-rank convolution layers, the communication cost savings reduces to $87.5\%$ to  $87.3\%$.} We observe, similarly to the results in \Cref{subsec:lin_regression}, that the maximum achieved communication cost savings, which depend on the layer ranks scales with the number of clients $C=4$, indicating that the decay rate of the singular values of the averaged coefficient matrix $\augS^*$ depends on $C$.

\begin{figure}
    \centering
\begin{subfigure}[t]{\textwidth}	
  \includegraphics[width=\textwidth]{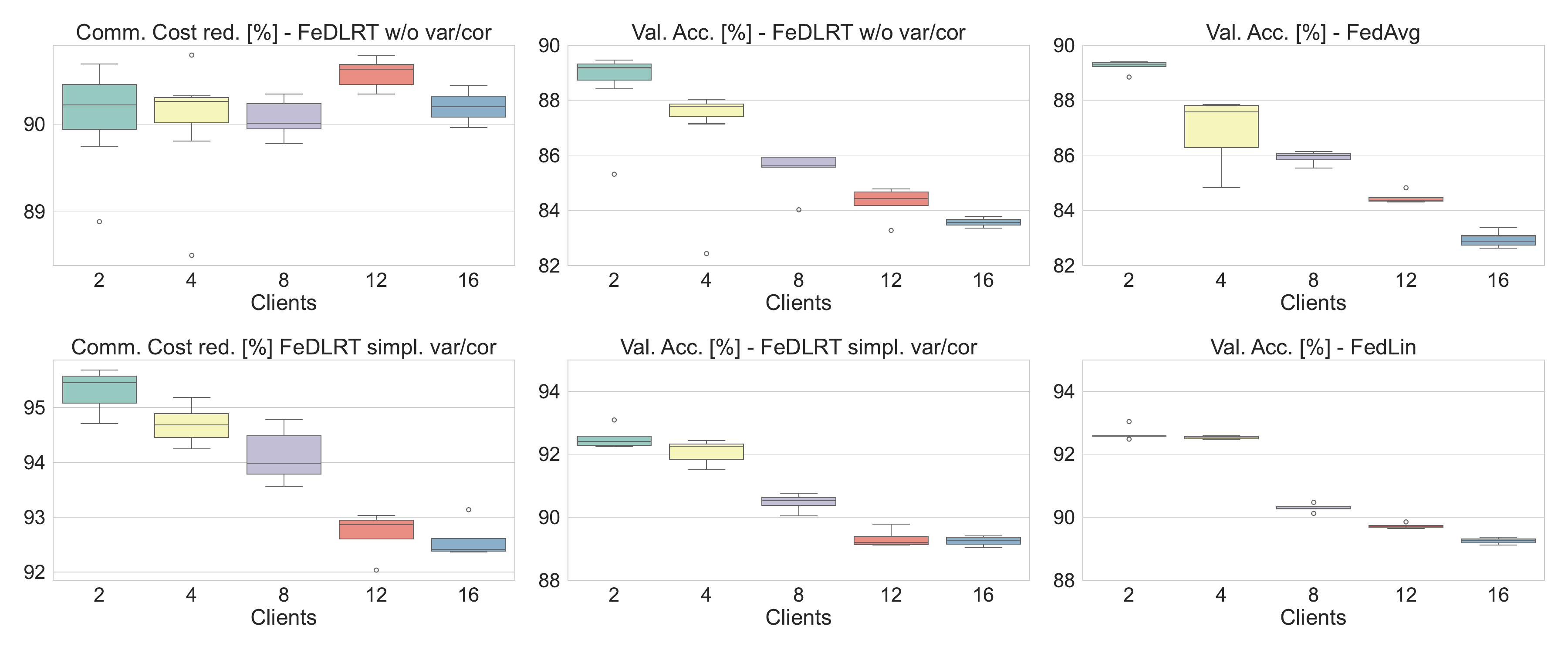}
    \end{subfigure}
    \caption{VGG16 CIFAR10 benchmark with $240/C$ local iterations for $C$ clients with  simplified (lower row) and without (upper row) variance correction. (Left panel) show the savings in communication cost corresponding to FedLin at final time. (Mid and right panel top row) compares the validation accuracy of FeDLRT and FedAvg, where we see that FeDLRT behaves similarly to FedAvg, where higher $C$ correlates with a drop in accuracy. FeDLRT with variance correction mitigates this issue and achieves similar performance as FedLin, close to the non-federated baseline accuracy is $93.15\%$. }
 \label{fig:vgg_cifar10}
\end{figure}

\begin{figure}
    \centering
\begin{subfigure}[t]{\textwidth}	
  \includegraphics[width=\textwidth]{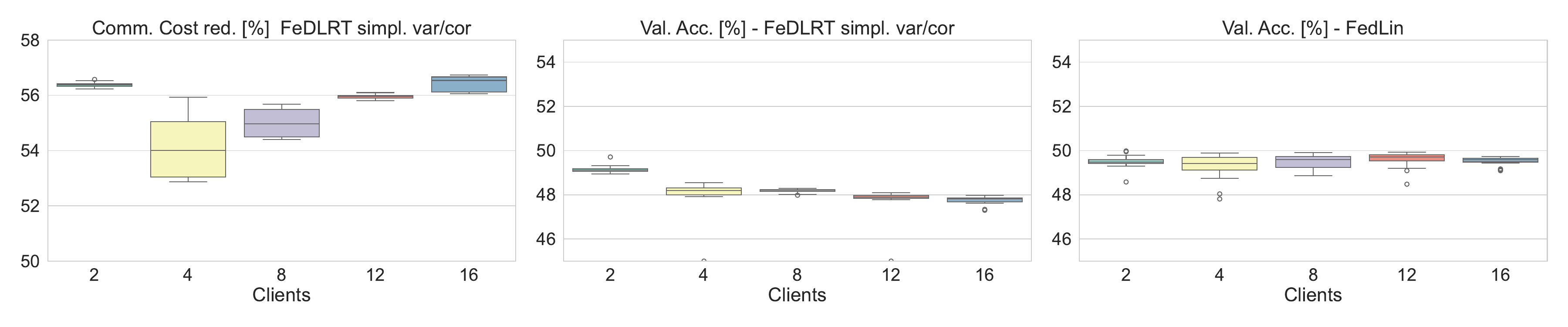}
    \end{subfigure}
    \caption{ViT CIFAR100 benchmark. (Left Panel) shows the savings in communication cost of variance corrected FeDLRT vs FedLin. (Mid and right panel) compares the validation accuracy of FeDLRT and FedLin, where we see that FeDLRT behaves similarly to FedLin and achieves accuracy levels near the non-federated baseline value of $50\%$, which is similar to literature results~\cite{zhu2023understanding}.   }
 \label{fig:vit_cifar100}
\end{figure}

\textbf{Vision Transformer on CIFAR100:} We consider a small vision transformer for CIFAR100, with $6$ attention layers with $2$ heads each followed by a ResNet block and a drop-out layer, all with weight matrices of dimension $512\times 512$. The tokenizer takes patches of size $8$ with embedding dimension $512$. Training hyperparameters are given in \Cref{tab:object_detection_setup}. Remark that we do not aim for SOTA performance, since transformer architectures are notoriously difficult to compress with low-rank approaches, but rather compare the performance of FedLin to FeDLRT for a given compute budget. We use $s_*=240/C$ local iterations for $C$ clients. Observe in \Cref{fig:vit_cifar100} that FeDLRT achieves similar performance as ViT with over $55\%$ communication cost savings on average.

\begin{table}[]
    \centering
        \caption{Experimental setup object detection benchmarks. All test cases use a cosine annealing learning rate scheduler. }

 \resizebox{\textwidth}{!}
 {
\begin{NiceTabular}{l| c c c c}
\toprule   
& {Alexnet/Cifar10}  &  {ResNet18/Cifar10} & VGG16/Cifar10 & {ViT/Cifar100}\\
\midrule
 Batch size & $128$ & $128$ & $128$ &$256$\\
 Start Learningrate &  $1\rm{e}{-2}$ & $1\rm{e}{-3}$ &  $1\rm{e}{-2}$ & $3\rm{e}{-4}$  \\
 End Learningrate &  $1\rm{e}{-5}$ &   $5\rm{e}{-4}$ & $5\rm{e}{-4}$ & $1\rm{e}{-5}$ \\
 Aggregation Rounds & $200$ &  $200$ &  $200$ &  $200$ \\
 Local Iterations & $100$ &  $240/C$ &$240/C$ &$240/C$ \\
 Truncation tolerance $\tau$ & $0.01$  & $0.01$ & $0.01$ & $0.01$ \\
 Momentum & $0.0$ &  $0.9$ &  $0.1$ &  n.a.   \\
 Weight Decay & $1\rm{e}{-4}$ & $1\rm{e}{-3}$  & $1\rm{e}{-4}$ &$1\rm{e}{-2}$  \\
 Optimizer & SGD & SGD & SGD &  Adam w/ std pytorch parameters \\
 \bottomrule
\end{NiceTabular}
}
    \label{tab:object_detection_setup}
\end{table}

\newpage
\section{Notation overview for the numerical analysis}
\label{appendix_notation}
We establish a set of notations to simplify the notation in the proofs
\begin{itemize}[leftmargin=*,noitemsep,topsep=0em]
\item $\mathcal{L}_c(W)$ denotes the local loss function based on dataset $X_c$ at client $c$.
\item $\mathcal{L}(W) = \frac{1}{C}\sum_{c=1}^C \mathcal{L}_c(W)$ is the global loss function.
      \item  $F_c(W)=-\nabla_W\mathcal{L}_c(W)$ is the negate of local loss gradient.
    \item  $F(W)= \frac{1}{C} \sum_{c=1}^C F_c(W)$ is the negate of global loss gradient.
    \item $\mathcal{M}_r = \set{W\in\mathbb{R}^{n \times n}: \text{rank}(W)=r}$ is a manifold of rank $r$ matrices.
    \item  $W_r =USV^\top\in \mathcal{M}_r$ is a rank-$r$ approximation of a matrix $W$.
    \item  $\mathcal{T}_{W_r} \mathcal{M}_{r}$ is the tangent space of $\mathcal{M}_r$ at $W_r$.
    \item  $P(W_r)$ is the orthogonal projection onto $\mathcal{T}_{W_r}\mathcal{M}_r$.
    \item  $P_U=UU^\top$ is the orthogonal projection onto the range of orthonormal $U\in\mathbb{R}^{n\times r}$.
    \item  $P_V=VV^\top$ is the orthogonal projection onto the range of orthonormal $V\in\mathbb{R}^{n\times r}$.
    \item When applied to vectors, $\norm{\cdot}$ denotes the Euclidean norm ($\ell_2$-norm). When applied to matrices, $\norm{\cdot}$ denotes the Frobenius norm.
\end{itemize}

\section{Efficient basis gradient dynamics for basis augmentation}\label{sec:grad_dynamics}
We first consider the basis update \& Galerkin splitting scheme of ~\eqref{eq:gradient_flow_system}. The splitting performs a reparametrization of the form  $K(t)=U(t)S(t)$ and $L(t)=V(t)S(t)^\top$. The basis update then reads 
\begin{align}\label{eq:system_ODEs_factors}
\begin{aligned}
\dot K = -\nabla_{K}\mathcal L(K(t)V_0^\top)\in\mathbb{R}^{n\times r}, \quad K(0) = U_0S_0, & \\
\dot L = -\nabla_{L}\mathcal L(U_0L(t)^\top)\in\mathbb{R}^{n\times r}, \quad L(0) = V_0S_0^\top.
\end{aligned}
\end{align}
Given the solution $K(t_1)$ and $L(t_1)$ at time $t_1$,  the bases $U_0$ and $V_0$ are augmented by the orthonormalization of the new directions $K(t_1)$ and $L(t_1)$, i.e.
\begin{align} \label{eq_basis_augmentation_KLs}
\begin{aligned}
\augU R & = \texttt{qr}([U_0 \mid K(t_1)])\in\mathbb{R}^{n\times 2r}, \\
\text{and}\quad \augV R &= \texttt{qr}([V_0 \mid L(t_1)])\in\mathbb{R}^{n\times 2r},
\end{aligned}
\end{align}
where $R$ is the right factor of the respective QR decomposition and can be discarded. 
The initial condition of the coefficient update is $S(t_0)$ projected onto the new bases, i.e., 
\begin{align}\label{eq:s_eq}
    \dot \augS = -\nabla_{S}\mathcal{L}(\augU \augS(t)\augV^\top), \quad \augS(0) = \augU^\top U_0 \augS(0) V_0^\top \augV. & 
\end{align}
After the integration of the coefficient dynamics above, the redundant basis functions are typically truncated via an SVD of $S$ ensuring that $S$ is always full rank. 
In its continuous form above, the splitting yields a robust integrator for the projected gradient flow, without manifold dependent step-size restrictions: 
\begin{theorem}{(\cite{Schothoefer_2022})}\label{robust_error_bound}
Assume $\mathcal{L}$ is $L$-smooth with constant $L$, and locally bounded by $B$. Let $\Wr(t)$ be the low-rank continuous time solution of \eqref{eq:system_ODEs_factors} and \eqref{eq:s_eq} and let $W(t)$ be the full rank solution at $t=0$. Assume the $K,L,$ and $S$ equations are integrated exactly from time $t=0$ to $\Delta t$. Assume that for any $Y\in\mathcal{M}_r$ sufficiently close to $\Wr(t)$ the gradient $F(Y)$ is $\epsilon$ close to $\mathcal{M}_r$. Then 
\begin{align*}
    \norm{W(\Delta t)-\Wr(\Delta t)}\leq d_1\epsilon + d_2 \Delta t + d_3\frac{\vartheta}{\Delta t},
\end{align*}
    where $d_1,d_2,d_3$ depend only on $L$ and $B$.
\end{theorem}
The theorem guarantees, that the low-rank representation does not imply any step-size restrictions on the optimization scheme. This is in stark contrast to a naive alternating descent optimization of the low-rank factors $U,S,V$. 

To build an discretized numerical optimizer in a resource constrained federated scenario from the above continuous splitting equations, we avoid the reparametrization, which implies a $200\%$ memory cost increase on the client side, since three versions of the low-rank layer need to be tracked.

\begin{lemma}\label{corr_gradient_trick}
Let $USV\in \mathcal M_r$ be a low rank factorization that follows the projected gradient \eqref{eq:gradient_flow_system} flow using the splitting scheme \eqref{eq:system_ODEs_factors} with $K=US$ and $V=VS^\top$. 
Further, assume that equations for the $K$ and $L$ factors are solved by an explicit Euler time integration with learning rate $\lambda$, i.e.
\begin{align}\label{eq:system_KL_euler}
\begin{aligned}
K(t_1) = K(0) -\lambda \nabla_{K}\mathcal L(K(0)V_0^\top), \quad K(0) = U_0S_0, & \\
L(t_1) = L(0) -\lambda\nabla_{L}\mathcal L(U_0L(0)^\top), \quad L(0) = V_0S_0^\top.
\end{aligned}
\end{align}

Then, the basis augmentation \eqref{eq_basis_augmentation_KLs} can be expressed as
\begin{align} 
\begin{aligned}
\augU R & =  \texttt{\textup{qr}}([U_0 \mid - \nabla_{U}\mathcal L\big(U_0S_0V_0^\top)])\in\mathbb{R}^{n\times 2r}, \\
\text{and}\quad \augV R & = \texttt{\textup{qr}}([V_0 \mid -\nabla_{V}\mathcal L\big(U_0S_0V_0^\top)])\in\mathbb{R}^{n\times 2r}.
\end{aligned}
\end{align}
and maintains the structure of the basis update and Galerkin operator split.
\end{lemma}
\begin{proof}
We consider the proof for the $K$ equation and the $U$ basis; the proof for $L$ and $V$ follows analogously.

Considering ~\eqref{eq_basis_augmentation_KLs}, we obtain with the explicit Euler discretization ~\eqref{eq:system_KL_euler}, 
\begin{align}
 \label{eq:grad_dynamics}
    \begin{aligned}
     \mathrm{span}\left([U_0 \mid K(t_1)]\right) &=   \mathrm{span}\left([U_0 \mid U_0 -\lambda \nabla_{K}\mathcal L(K(0)V_0^\top)]\right)  \\
     &=   \mathrm{span}\left([U_0 \mid -\lambda \nabla_{K}\mathcal L(K(0)V_0^\top)]\right)  \\
       &=   \mathrm{span}\left([U_0 \mid -\nabla_{K}\mathcal L(K(0)V_0^\top)]\right).
    \end{aligned}
\end{align}
Next, consider the continuous time dynamics of $\dot K$, where we omit explicit time dependence on $U,S,V$ and $K$ for the sake of brevity, i.e.,
\begin{align} \label{eq:Ki_reduced}
    \begin{aligned}
    \dot{K}
        &= \dot{\left(US\right)} \\ 
        &= \dot{U} S + U \dot{S} \\
        &  \overset{\text{~\eqref{eq:gradient_flow_system}}}{=}
        -(I-U U^{\top})\nabla_{W}\mathcal{L}(USV^\top)V S^{-1}S - U U^{\top} \nabla_{W}\mathcal{L}(USV^\top) V  \\
        &= - (I -P_U)\nabla_{W}\mathcal{L}(USV^\top)V - P_U\nabla_{W}\mathcal{L}(USV^\top) V \\
       &= (P_U- I)\nabla_{W}\mathcal{L}(USV^\top)V - P_U\nabla_{W}\mathcal{L}(USV^\top) V \\
       &=-\nabla_{W}\mathcal{L}(USV^\top)V
    \end{aligned}
\end{align}
Further, using the chain rule, we observe
\begin{align*}
        \nabla_{U}\mathcal{L}(USV^\top) =\nabla_W\mathcal{L}(USV^\top)\nabla_{U}(USV^\top)=\nabla_W\mathcal{L}(USV^\top)VS^\top
    \end{align*}
Thus, $-\nabla_{U}\mathcal{L}(USV^\top)S^{-\top} = -\nabla_W\mathcal{L}(USV^\top)V= \dot K$. 
Full rankness of $S$  and ~\eqref{eq:Ki_reduced} yield that $
     \mathrm{span}(-\nabla_{U}\mathcal{L}(USV^\top)) = \mathrm{span}(\dot K)$.
     Together with ~\eqref{eq:grad_dynamics} this yields the proof.
\end{proof}
\Cref{corr_gradient_trick} adopts a more general result for Tucker tensors in an unpublished manuscript and simplifies the analysis for the matrix case considered here.

\section{Efficient basis and coefficient communication}\label{sec:basis_com}
Note that we have by orthogonality of the bases $\augU=[U,\Bar{U}]$ with $\Bar{U}\in\mathbb{R}^{n\times r}$ and $\Bar{U}^\top U =0$ and $\augV=[V,\Bar{V}]$ with $\Bar{V}\in\mathbb{R}^{n\times r}$ and $\Bar{V}^\top V =0$.

\begin{proof}{(\Cref{lem_aug_coeff})}
The basis augmented basis $[U,G_U]$ before orthonormalization already contains the orthonormal vectors given by the columns of $U$. A QR decomposition therefor only rearranges the columns of $G_U$ such that $\augU=[U,\Bar{U}]$ with $\Bar{U}\in\mathbb{R}^{n\times r}$ and $\Bar{U}^\top U =0$. The analogous result holds for   $\augV=[V,\Bar{V}]$. The projection onto the augmented basis therefore reads
\begin{align}
  \augU^\top U =
\begin{bmatrix}
{U}^\top U \\
\overline{U}^\top U
\end{bmatrix}
=
\begin{bmatrix}
I \\
0
\end{bmatrix}
\quad\text{and} \quad     
\augV^\top V =
\begin{bmatrix}
{V}^\top V \\
\overline{V}^\top V
\end{bmatrix}
=
\begin{bmatrix}
I \\
0
\end{bmatrix}.
\end{align}
Consequently, the augmented coefficient matrix takes the form
\begin{align}
    \augS= \augU^\top U S V^\top \augV =
\begin{bmatrix}
S & 0 \\
0 & 0
\end{bmatrix}.
\end{align}
\end{proof}

\section{Analysis for FeDLRT with full variance correction}
In this section we establish bounds on the coefficient drift of the FeDLRT method with full variance correction.  We use the established coefficient drift bound to derive a loss-descend guarantee. The strategy of our analysis follows the one of FedLin~\cite{NEURIPS2021_7a6bda9a}.
We first state an auxiliary lemma.
\begin{lemma}\label{lem_helper_projectedLipschitz}
Let $U\in\mathbb{R}^{n\times r}$ and $V\in\mathbb{R}^{n\times r}$ be orthonormal matrices. 
Let $F$ be an $L$-continuous function.
Then, for $S_1,S_2\in\mathbb{R}^{r\times r}$,
    \begin{align}
        \norm{P_U\left(F(US_1V^\top)-F(US_2V^\top)\right) P_V}\leq L\norm{S_1 -S_2}
    \end{align}
    and 
       \begin{align}
        \norm{U\left(F(US_1V^\top)-F(US_2V^\top)\right) V^\top}\leq L\norm{S_1 -S_2},
    \end{align}
where $P_U$ and $P_V$ are orthogonal projections defined in \Cref{appendix_notation}.
\end{lemma}
\begin{proof}
    For the first statement, consider
    \begin{align*}
                &\norm{P_U\left(F(US_1V^\top)-F(US_2V^\top)\right) P_V} \\
                =&\norm{UU^\top\left(F(US_1V^\top)-F(US_2V^\top)\right) VV^\top}\\
                   \overset{\text{(I)}}{\leq} &\norm{U}\norm{U^\top}\norm{F(US_1V^\top)-F(US_2V^\top)} \norm{V}\norm{V^\top} \\
                 \overset{\text{(II)}}{=} &\norm{F(US_1V^\top)-F(US_2V^\top)} \\
                   \overset{\text{(III)}}{\leq} &L \norm{US_1V^\top-US_2V^\top} =L \norm{U(S_1-S_2)V^\top}\\
                     \overset{\text{(I)}}{\leq} &L \norm{U}\norm{S_1-S_2}\norm{V^\top} \\
                        \overset{\text{(II)}}{=} &L \norm{S_1-S_2},
    \end{align*}
where we have used in (I) the operator norm inequality of the Frobenius norm, in (II) orthonormality of $U$, $V$, and in (III) $L$-continuity of $F$. The second statement is proven analogously.
\end{proof}

\subsection{Coefficient drift bound for FeDLRT with full variance correction} \label{sec:var_reduction_FeDLRT_full}
We consider the FeDLRT method with variance correction, see \Cref{alg:FeDLRT_variance_reduction}. Key difference to the FeDLRT method without variance correction is the modified coefficient update, incorporating global gradient information of the augmented coefficient matrix $\augS$ and local, stale gradient information of the augmented coefficient matrix $\augS_c$. 
The variance corrected local coefficient update \eqref{eq_grad_update_var_full} can be expressed in terms of the projected Riemannian gradient as
\begin{align}\label{eq_new_varcor_coeff_update_full}
    \augSc^{s+1} = \augSc^s +  \lambda \augU^\top\left( F_c(\augWrc^s) - F_c(\augWr) + F(\augWr)\right)\augV,
\end{align}
where $\augU^\top F_c(\augWrc^s)\augV = \nabla_{\augSc}\mathcal{L}_c(\augU\augSc^s\augV) $, $\augU^\top F_c(\augWrc)\augV = \nabla_{\augSc}\mathcal{L}_c(\augU\augSc^{s=0}\augV) $ and $\augU^\top F_c(\augWrc^s)\augV = \nabla_{\augSc}\mathcal{L}(\augU\augSc^s\augV) $. Recall that $ \augS=\augSc$ for $s=0$.

We provide proof for Theorem~\ref{lem_drift_bound_full} to bound the drift term $\norm{\augSc^s-\augSc}$. We restate this theorem to the Riemannian notation and restate it below.
\begin{theorem}{(Restatement of Theorem~\ref{lem_drift_bound_full})}\label{lem_drift_bound_full_translated}
Given augmented basis and coefficient matrices $\augU$, $\augV$, and $\augS$, and $\augWr=\augU\augS\augV^\top$. 
If the local learning rate $0<\lambda\leq\frac{1}{Ls_*}$ with $s_*\geq 1$ the number of local steps, for all clients $c$,
    \begin{align}
    \textstyle
        \|{\augSc^s-\augSc}\|\leq \exp(1)s_* \lambda\norm{ \augU^\top F(\augWr)\augV}, \quad\textup{for}\quad s=1,\dots,s^*-1,
    \end{align}
where $\augSc^s$ is the variance corrected coefficient as given in \eqref{eq_grad_update_var_full}.
\end{theorem}
\begin{proof}
    From the adjusted coefficient update in \eqref{eq_new_varcor_coeff_update_full}, we get
    \begin{align*}
    \norm{\augSc^{s+1} -\augSc}
    &=\norm{\augSc^s -\augSc + \lambda \augU^\top\left( F_c(\augWrc^s) - F_c(\augWr) + F(\augWr)\right)\augV}\\
    &\leq \norm{\augSc^s -\augSc} + \lambda \norm{\augU^\top\left( F_c(\augWrc^s) - F_c(\augWr)\right)\augV} + \lambda\norm{ \augU^\top F(\augWr)\augV}\\
      & \overset{\text{(I)}}{\leq} \norm{\augSc^s -\augSc} + \lambda L \norm{\augSc^s - \augS} + \lambda\norm{ \augU^\top F(\augWr)\augV}\\
&\leq (1 + \lambda L) \norm{\augSc^s - \augS} + \lambda\norm{ \augU^\top F(\augWr)\augV}\\
&\leq \left(1 + \frac{1}{s_*}\right) \norm{\augSc^s - \augS} + \lambda\norm{ \augU^\top F(\augWr)\augV}.
\end{align*}
We use in (I) Lemma~\ref{lem_helper_projectedLipschitz} 
Recursively plugging in the above inequality yields for $a=(1 +  \frac{1}{s_*})$
\begin{align*}
     \norm{\augSc^{s+1} -\augSc}
    &\leq a^{s+1} \norm{\augSc^{s=0} - \augS} + \left(\sum_{j=0}^s  a^{j}\right) \lambda\norm{ \augU^\top F(\augWr)\augV}\\
    &=  \left(\sum_{j=0}^s  a^{j}\right) \lambda\norm{ \augU^\top F(\augWr)\augV}\\
     &=   \frac{ a^{s+1} -1 }{a -1} \lambda\norm{ \augU^\top F(\augWr)\augV}\\
     &\leq  \left(1+\frac{1}{s_*}\right)^{s+1} s_* \lambda\norm{ \augU^\top F(\augWr)\augV}\\
     &\leq  \left(1+\frac{1}{s_*}\right)^{ s_*} s_* \lambda\norm{ \augU^\top F(\augWr)\augV}\\
     &\leq  \exp(1) s_* \lambda\norm{ \augU^\top F(\augWr)\augV}.
\end{align*}
\end{proof}

\subsection{Global loss descent for FeDLRT with full variance correction}\label{sec_app_proof_theo_bound_deterministic_variance_reduction_full}

We first state a few auxiliary lemmas, which provide common inequalities that will be used in the following analysis.
\begin{lemma}{(\cite[Lemma 5.2]{HnatiukPaper})}\label{lem_loss_function_est}
    For any two matrices $Y_1,Y_2\in\mathbb{R}^{n\times n}$ and an $L$-smooth $\mathcal{L}$ with constant $L$ it holds 
    \begin{align}
        \mathcal{L}(Y_1)-\mathcal{L}(Y_2)\leq -\inner{Y_1-Y_2,F(Y_2)} + \frac{L}{2}\norm{Y_1-Y_2}^2,
    \end{align}
    where $F(Y)=-\nabla_Y\mathcal{L}(Y)$.
\end{lemma}
\begin{lemma}{(\cite[Lemma 5]{mitra2021linear})}\label{lem_helper1}
    For two vectors $x_1,x_2\in\mathbb{R}^d$ it holds for $\gamma>0$
    \begin{align}
      \norm{x_1+x_2}^2\leq(1+\gamma)\norm{x_1}^2 + \left(1+\frac{1}{\gamma}\right)\norm{x_2}^2.
    \end{align}
\end{lemma}
   \begin{lemma}{(\cite[Lemma 6]{mitra2021linear})}\label{lem_helper2}
    For $C$ vectors $x_1,\dots,x_C\in\mathbb{R}^d$ the application of Jensen's inequality yields
    \begin{align}
        \norm{\sum_{c=1}^C x_c}^2\leq C \sum_{c=1}^C\norm{x_c}^2.
    \end{align}
\end{lemma} 
First, we consider the loss function value at the augmentation step.
\begin{lemma}\label{lem_loss_aug}
     We have $\mathcal{L}(\augWr)=\mathcal{L}(W_r^t)$ for the loss before and after basis augmentation. 
\end{lemma}
\begin{proof}
    Due to \Cref{lem_aug_coeff}, $\augS =
\begin{bmatrix}
S^t & 0 \\
0 & 0
\end{bmatrix}$, thus $\augWr=\augU\augS\augV^\top = USV^\top = W^t$.
\end{proof}
We next bound the loss descent between the augmentation step and the truncation step - having performed the aggregation of the client updates.
\begin{theorem}\label{theo_client_descent_bound}Let $\augWr=\augU \augS \augV^{\top}$ be the augmented factorization at global iteration $t$ and let
$ \augWr^*=\augU \augS^* \augV^{\top}$ be the aggregated solution after client iterations, i.e., $\augS^{*} =  \frac{1}{C} \sum_{c=1}^C\augS_c^{s_*}$.
Then the variance corrected coefficient update \eqref{eq_new_varcor_coeff_update_full} yields the guarantee
\begin{equation}\label{eq_theo_client_descent_bound}
\begin{alignedat}{2}
    \mathcal{L}(\augWr^*)-\mathcal{L}(\augWr) &\leq -( s_*\lambda)(1-( s_*\lambda)L)\norm{\augU^\top F(\augWr)\augV}^2\\
    &+ \left(\frac{L\lambda}{C}\sum_{c=1}^C\sum_{s=0}^{s_*-1}\norm{ \augSc^s -\augS} \right)\norm{\augU^\top F(\augWr)\augV} \\
    &+ \frac{L^3\lambda^2 s_*}{C}\sum_{c=1}^C \sum_{s=0}^{s_*-1} \norm{\augSc^s-\augSc}^2.
\end{alignedat}
\end{equation}
\end{theorem}
\begin{proof}
  From \eqref{eq_grad_update_var_full}, $P_\augU =\augU\augU^\top$, $P_\augV =\augV\augV^\top$, and the fact that $\augWrc^{s=0}=\augWr$ for all $c=1,\dots,C$,
  \begin{align*}
      \augWrc^{s_*} &= \augU\augS_c^{s_*}\augV^\top = \augU\augS_c^{s=0}\augV^\top +  \augU\augU^\top\sum_{s=0}^{s_*-1} \lambda\left( F_c(\augWrc^s) - F_c(\augWr) + F(\augWr)\right)\augV\augV^\top \\
     & = \augWr -\lambda\sum_{s=0}^{s_*-1}P_\augU F_c(\augWrc^s)P_\augV - \lambda P_\augU\left(F(\augWr)-F_c(\augWr)\right)P_\augV.
\end{align*}
Averaging across clients leads to
\begin{align}
    \augWr^* =& \frac{1}{C}\sum_{c=1}^C\augWrc^{s_*} = \augWr - \frac{\lambda}{C}\sum_{c=1}^C\sum_{s=0}^{s_*-1}  P_\augU F_c(\augWrc^s)  P_\augV-  \frac{\lambda}{C}\sum_{c=1}^C
P_\augU\left(F(\augWr)-F_c(\augWr)\right)  P_\augV \nonumber\\
=&  \augWr - \frac{\lambda}{C}\sum_{c=1}^C\sum_{s=0}^{s_*-1}  P_\augU F_c(\augWrc^s)  P_\augV,
\label{eq_helper_1_1}
\end{align}
where we have used the definition of the global and local gradient at $\augWr$, i.e., $\frac{1}{C}\sum_{c=1}^CF_c(\augWr) = F(\augWr)$. Based on $L$-continuity of $F$ and $F_c$, \eqref{eq_helper_1_1}, and \Cref{lem_loss_function_est}, we obtain further
\begin{align}\label{eq_loss_descent_ineq}
    \mathcal{L}(\augWr^*)-\mathcal{L}(\augWr)&\leq \inner{\augWr^*-\augWr,F(\augWr)} + \frac{L}{2}\norm{\augWr^*-\augWr}^2\\
    &= -\inner{ \frac{\lambda}{C}\sum_{c=1}^C\sum_{s=0}^{s_*-1} P_\augU F_c(\augWrc^s)P_\augV ,F(\augWr)} +\frac{L}{2}\norm{ \frac{\lambda}{C}\sum_{c=1}^C\sum_{s=0}^{s_*-1} P_\augU F_c(\augWrc^s) P_\augV}^2.\nonumber
\end{align}
Next, we bound each of the two right-hand-side terms separately. We first express the first term as
\begin{align*}
    &-\inner{ \frac{\lambda}{C}\sum_{c=1}^C\sum_{s=0}^{s_*-1}  P_\augU F_c(\augWrc^s)  P_\augV,F(\augWr)} \\
    =& -\inner{ \frac{\lambda}{C}\sum_{c=1}^C\sum_{s=0}^{s_*-1} P_\augU \left( F_c(\augWrc^s) -F_c(\augWr)\right)P_\augV + P_\augU \left(\frac{\lambda}{C}\sum_{c=1}^C\sum_{s=0}^{s_*-1} F_c(\augWr)\right)P_\augV ,F(\augWr)} \\
    =&-\inner{ \frac{\lambda}{C}\sum_{c=1}^C\sum_{s=0}^{s_*-1} P_\augU \left( F_c(\augWrc^s) -F_c(\augWr)\right)P_\augV +  P_\augU  \frac{s_*\lambda}{C}\sum_{c=1}^C F_c(\augWr) P_\augV ,F(\augWr)} \\
    =&-\inner{ P_\augU \left(\frac{\lambda}{C}\sum_{c=1}^C\sum_{s=0}^{s_*-1} F_c(\augWrc^s) -F_c(\augWr)\right)P_\augV +P_\augU s_*\lambda F(\augWr) P_\augV,F(\augWr)} \\
    =&-\inner{ \augU^\top\left( \frac{\lambda}{C}\sum_{c=1}^C\sum_{s=0}^{s_*-1} F_c(\augWrc^s) -F_c(\augWr)\right)\augV  , \augU^\top F(\augWr) \augV^\top}  -  s_*\lambda \inner { \augU^\top F(\augWr)\augV,\augU^\top F(\augWr)\augV }\\
       =&-\inner{ \frac{\lambda}{C}\sum_{c=1}^C\sum_{s=0}^{s_*-1} \augU^\top \left( F_c(\augWrc^s) -F_c(\augWr)\right)\augV  , \augU^\top F(\augWr) \augV}  -  s_*\lambda \norm{\augU^\top F(\augWr)\augV}^2,
\end{align*}
where the definitions of $P_\augU$ and $P_\augV$ are used.
Following this, the first term then can be bounded by
\begin{align*}
&-\inner{ \frac{\lambda}{C}\sum_{c=1}^C\sum_{s=0}^{s_*-1}  P_\augU F_c(\augWrc^s)  P_\augV,F(\augWr)} \\
{\leq} & \frac{\lambda}{C}\sum_{c=1}^C\sum_{s=0}^{s_*-1}\norm{ \augU^\top  \left( F_c(\augWrc^s) -F_c(\augWr)\right)\augV}\norm{ \augU ^\top  F(\augWr) \augV}  -  s_*\lambda \norm{\augU^\top F(\augWr)\augV}^2\\
{\leq} & \frac{L\lambda}{C}\sum_{c=1}^C\sum_{s=0}^{s_*-1}\norm{ \augSc^s -\augS}\norm{\augU^\top F(\augWr)\augV}  -  s_*\lambda \norm{\augU^\top F(\augWr)\augV}^2,
\end{align*}
where Lemma~\ref{lem_helper_projectedLipschitz} is invoked in the last inequality.
Following a similar approach, we express the second term as
\begin{align*}
    &\frac{L}{2}\norm{ \frac{\lambda}{C}\sum_{c=1}^C\sum_{s=0}^{s_*-1} P_\augU  F_c(\augWrc^s) P_\augV }^2
    =\frac{L}{2}\norm{ \frac{\lambda}{C}\sum_{c=1}^C\sum_{s=0}^{s_*-1}P_\augU \left( F_c(\augWrc^s)-F_c(\augWr)\right)P_\augV+ s_*\lambda P_\augU F(\augWr)P_\augV}^2,
    \end{align*}
which can be bounded by
    \begin{align*}
    &\frac{L}{2}\norm{ \frac{\lambda}{C}\sum_{c=1}^C\sum_{s=0}^{s_*-1} P_\augU  F_c(\augWrc^s) P_\augV }^2\\
      \overset{\text{(I)}}{\leq} & L\norm{ \frac{\lambda}{C}\sum_{c=1}^C\sum_{s=0}^{s_*-1}P_\augU \left( F_c(\augWrc^s)-F_c(\augWr)\right)P_\augV}^2+( s_*\lambda)^2L\norm{P_\augU F(\augWr)P_\augV}^2 \\
          \overset{\text{(II)}}{\leq} &  \frac{L}{C}\sum_{c=1}^C\lambda^2 s_* \sum_{s=0}^{s_*-1} \norm{P_\augU \left( F_c(\augWrc^s)-F_c(\augWr)\right)P_\augV}^2+( s_*\lambda)^2L\norm{P_\augU F(\augWr)P_\augV}^2\\
          \overset{\text{(III)}}{\leq} &  \frac{L^3\lambda^2 s_*}{C}\sum_{c=1}^C \sum_{s=0}^{s_*-1}  \norm{\augSc^s-\augSc}^2+( s_*\lambda)^2L\norm{ P_\augU F(\augWr)P_\augV}^2\\
          \overset{\text{(IV)}}{\leq} &  \frac{L^3\lambda^2 s_*}{C}\sum_{c=1}^C \sum_{s=0}^{s_*-1} \norm{\augSc^s-\augSc}^2+( s_*\lambda)^2L\norm{\augU^\top F(\augWr)\augV}^2,
\end{align*}
where Lemma~\ref{lem_helper1} with $\gamma=1$ is used in in (I), \text{Jensen's inequality} is used in (II),  \text{Lemma~\ref{lem_helper_projectedLipschitz}} is used in in (III), and (IV) follows from the Operator norm inequality of the Frobenius norm in combination with orthonormality of $U$ and $V^\top$.

Plugging these two bounds into \eqref{eq_loss_descent_ineq} gives
\begin{align*}
     \mathcal{L}(\augWr^*)-\mathcal{L}(\augWr)
      \leq& -\inner{ \frac{\lambda}{C}\sum_{c=1}^C\sum_{s=0}^{s_*-1} P_\augU F_c(\augWrc^s)P_\augV ,F(\augWr)} +\frac{L}{2}\norm{ \frac{\lambda}{C}\sum_{c=1}^C\sum_{s=0}^{s_*-1} P_\augU F_c(\augWrc^s) P_\augV}^2 \\
    \leq& \frac{L\lambda}{C}\sum_{c=1}^C\sum_{s=0}^{s_*-1}\norm{ \augSc^s -\augS}\norm{\augU^\top F(\augWr)\augV}  -  s_*\lambda \norm{\augU^\top F(\augWr)\augV}^2   \\
    & + \frac{L^3\lambda^2 s_*}{C}\sum_{c=1}^C \sum_{s=0}^{s_*-1} \norm{\augSc^s-\augSc}^2+( s_*\lambda)^2L\norm{\augU^\top F(\augWr)\augV}^2 \\
    =& -( s_*\lambda)(1-( s_*\lambda)L)\norm{\augU^\top F(\augWr)\augV}^2\\
    &+ \left(\frac{L\lambda}{C}\sum_{c=1}^C\sum_{s=0}^{s_*-1}\norm{ \augSc^s -\augS} \right)\norm{\augU^\top F(\augWr)\augV}\\
    &+ \frac{L^3\lambda^2 s_*}{C}\sum_{c=1}^C \sum_{s=0}^{s_*-1} \norm{\augSc^s-\augSc}^2,
\end{align*}
which concludes the proof.
\end{proof}
With this result, we next bound the loss descent between the augmentation and coefficient aggregation step in the following theorem.
\begin{theorem}\label{theo_client_descent_bound2} 
Under the same assumptions as in \Cref{theo_client_descent_bound}.
Let the local learning rate be $0<\lambda\leq\frac{1}{12 Ls_*}$ with number of local iterations $s_*\geq 1$. Then,
\begin{align}
    \mathcal{L}(\augWr^*)-\mathcal{L}(\augWr)&\leq - s_*\lambda(1- 12 s_*\lambda L) \norm{\augU^\top F(\augWr)\augV}^2.
\end{align}
\end{theorem}
\begin{proof}

Applying the drift bound given in \Cref{lem_drift_bound_full} to the loss descent bound given by \Cref{theo_client_descent_bound} in \eqref{eq_theo_client_descent_bound} leads to
\begin{align*}
    & -( s_*\lambda)(1-( s_*\lambda)L)\norm{\augU^\top F(\augWr)\augV}^2\\
    &+ \left(\frac{L\lambda}{C}\sum_{c=1}^C\sum_{s=0}^{s_*-1}\left(\exp(1)s_* \lambda\norm{ \augU^\top F(\augWr)\augV}\right) \right)\norm{\augU^\top F(\augWr)\augV} \\
    & + \frac{L^3\lambda^2 s_*}{C}\sum_{c=1}^C \sum_{s=0}^{s_*-1} \left(\exp(1)s_* \lambda\norm{ \augU^\top F(\augWr)\augV}\right)^2 \\
    =& -( s_*\lambda)(1-( s_*\lambda)L)\norm{\augU^\top F(\augWr)\augV}^2 +L\lambda^2 s_*^2 \exp(1)\norm{ \augU^\top F(\augWr)\augV}^2\\
    &+ {L^3\lambda^4 s_*^4\exp(2)}\norm{ \augU^\top F(\augWr)\augV}^2\\
    =& -( s_*\lambda)(1-( s_*\lambda)L - ( s_*\lambda)L\exp(1) - ( s_*\lambda)^3 L^2 \exp(2) ) \norm{\augU^\top F(\augWr)\augV}^2\\
    \leq & -( s_*\lambda)(1-( s_*\lambda)L (1+\exp(1)+\exp(2))) \norm{\augU^\top F(\augWr)\augV}^2\\
    \leq & -( s_*\lambda)(1- 12( s_*\lambda)L) \norm{\augU^\top F(\augWr)\augV}^2,
\end{align*}
where we have used that $( s_*\lambda)L\leq 1$ and that $1+\exp(1)+\exp(2)\approx 11.107\leq 12$.    
\end{proof}

We are now prepared to prove \Cref{theo_loss_descent_full_var_cor}, which we restate in terms of Riemannian gradients as below.
\begin{theorem}{(Restatement of \Cref{theo_loss_descent_full_var_cor})}\label{theo_loss_descent_full_var_cor_translated}
Let $U^t S^t V^{t,\top}$  and $U^{t+1} S^{t+1} V^{t+1,\top}$ be the factorization before and after iteration $t$ of 
\Cref{alg:FeDLRT_variance_reduction} with variance correction and singular value truncation threshold $\vartheta$. Let $\mathcal{L}_c$ and $\mathcal{L}$ be $L$-smooth with constant $L$, and let the local learning rate be $0\leq\lambda\leq\frac{1}{12 Ls_*}$. Then the global loss descent is bounded by
\begin{align}
          \mathcal{L}(U^{t+1} S^{t+1} V^{t+1,\top}) - \mathcal{L}(U^t S^t V^{t,\top}) \leq -( s_*\lambda)(1- 12( s_*\lambda)L) \norm{\augU^\top F(\augWr)\augV}^2 + L\vartheta.
\end{align}
\end{theorem}

\begin{proof}
    Consider $\mathcal{L}(W_r^{t+1})$ and $\mathcal{L}(\augWr^{*})$, i.e., the loss values before and after the truncation step.
   By the mean value theorem, we obtain for some $h\in[0,1]$
   \begin{align}\label{eq_truncation_helperxxx}
   \begin{aligned}
       \mathcal{L}(W_r^{t+1}) &=  \mathcal{L}(\augWr^*) + \inner{-F(h W_r^{t+1} +(1-h)\augWr^*),W_r^{t+1} - \augWr^*}\\
       &{\leq}  \mathcal{L}(\augWr^*) + \norm{F(h W_r^{t+1} +(1-h)\augWr^*)}\norm{W_r^{t+1} - \augWr^*}\\
       &{\leq} \mathcal{L}(\augWr^*) + L\vartheta
   \end{aligned}
   \end{align}
   where $L$-smoothness and the fact that $\vartheta \geq \norm{W_r^{t+1} - \augWr^*}$ are used in (II), where the latter follows from the singular value truncation threshold.
   Combining the above arguments with \Cref{lem_loss_aug} and \Cref{theo_client_descent_bound2}  yields
\begin{align*}
       \mathcal{L}(W_r^{t+1}) - \mathcal{L}(W_r^{t}) &=  (\mathcal{L}(W_r^{t+1}) - \mathcal{L}(\augWr^{*})) +  (\mathcal{L}(\augWr^{*}) -  \mathcal{L}(\augWr)) +  (\mathcal{L}(\augWr) - \mathcal{L}(W_r^{t})) \\
      &\leq  L\vartheta  -( s_*\lambda)(1- 12( s_*\lambda)L) \norm{\augU^\top F(\augWr)\augV}^2,
\end{align*}
  which concludes the proof.
\end{proof}

\subsection{Global convergence of FeDLRT with full variance correction}\label{sub_global_convergence}
\begin{theorem}{(Restatement of \Cref{theo_glob_convergence})}
     Assume that $\mathcal{L}$ is $L$-smooth with constant $L$ for all $c=1,\dots,C$. Let $\augU^t \augS^t \augV^{t,\top}$ be the augmented representation at iteration $t$. Then Algorithm ~\ref{alg:FeDLRT_variance_reduction} guarantees for the learning rate $\lambda\leq\frac{1}{12 Ls_*}$ and final iteration $T$
    \begin{align}
          \min_{t=1,\dots,T}\norm{\nabla_\augS\mathcal{L}(U^t S^t V^{t,\top})}^2\leq \frac{48 L}{T}\left(\mathcal{L}(\Wr^{t=1})-\mathcal{L}(\Wr^{t={T+1}})\right) + {48 L^2}{\vartheta}.
    \end{align}
\end{theorem}
\begin{proof}

Consider \Cref{theo_loss_descent_full_var_cor}, 
\begin{align}\label{eq_helper123097213}
       \mathcal{L}(W_r^{t+1}) - \mathcal{L}(W_r^{t})       &\leq  L\vartheta  -( s_*\lambda)(1- 12( s_*\lambda)L) \norm{\nabla_\augS\mathcal{L}(U^t S^t V^{t,\top})}^2,
\end{align}
and assume that $\lambda s_* = \frac{1}{24L}$, i.e. $\lambda=\frac{1}{24L s_*}\leq\frac{1}{Ls_*}$, which obeys the learning rate requirement of Theorem~\ref{theo_loss_descent_full_var_cor}. Plugging this learning rate into \eqref{eq_helper123097213} gives
\begin{align*}
          \norm{\nabla_\augS\mathcal{L}(U^t S^t V^{t,\top})}^2\leq 48 L \left(\mathcal{L}(\Wr^t)-\mathcal{L}(\Wr^{t+1})+ L\vartheta\right).
\end{align*}
Averaging from $t=1$ to $t=T$ yields
\begin{align*}
     \min_{t=1,\dots,T}\norm{\nabla_\augS\mathcal{L}(U^t S^t V^{t,\top})}^2&\leq\frac{1}{T}\sum_{t=1}^T\norm{\nabla_\augS\mathcal{L}(U^t S^t V^{t,\top})}^2\\
     &\leq \frac{48 L}{T}\left(\mathcal{L}(\Wr^{t=1})-\mathcal{L}(\Wr^{t={T+1}})\right) + {48 L^2}{\vartheta},
\end{align*}
which concludes the proof.
\end{proof}

\section{Analysis for FeDLRT with simplified variance correction}

\label{sec:mod_var_reduction_FeDLRT}
We consider the FeDLRT method with simplified variance correction, see \Cref{alg:modified_FeDLRT_var_cor}.
Key difference to the standard FeDLRT with full variance correction, see \Cref{alg:FeDLRT_variance_reduction} is the modified coefficient update, incorporating global gradient information of the non-augmented coefficient matrix $S$ for the variance correction term, that is 
\begin{align}
\textstyle
   \check{V}_c = \check{G}_\augS - \check{G}_{\augS,c} = 
\begin{bmatrix}
\nabla_S \mathcal{L}(U^tS^tV^{t,\top}) - \nabla_S \mathcal{L}_c(U^tS^tV^{t,\top}) & 0 \\
0 & 0
\end{bmatrix}.
\end{align}
Using the Riemmanian gradient, we can equivalently write
\begin{align*}
    \check{V}_c = 
\left[U^\top|\,0 \,\right]( 
F(\augWr)-F_c(\augWr))\begin{bmatrix}
V \\
0 
\end{bmatrix}
= \augU^\top\begin{bmatrix}
I & 0 \\
0 &0 
\end{bmatrix}(F_c(\augWr) -
F(\augWr))\begin{bmatrix}
I & 0 \\
0 &0 
\end{bmatrix}\augV.
\end{align*}
 Remember the simplified variance corrected local coefficient update, given by
\begin{align}\label{eq_new_varcor_coeff_update}
\begin{aligned}
    \augSc^{s+1} &= \augSc^s +  \lambda \augU^\top\left( F_c(\augWrc^s) +\begin{bmatrix}
I & 0 \\
0 &0 
\end{bmatrix}(F_C(\augWr) -
F(\augWr))\begin{bmatrix}
I & 0 \\
0 &0 
\end{bmatrix}\right)\augV \\
&= \augSc^s +  \lambda \augU^\top\left( F_c(\augWrc^s) \right)\augV +\check{V}_c.
\end{aligned}
\end{align}

\subsection{Global loss descent for FeDLRT with simplified variance correction}\label{sec_app_proof_theo_bound_deterministic_simple_variance_reduction}
In the following we provide proof for a global loss descent for \Cref{alg:modified_FeDLRT_var_cor}, i.e. using the 
local coefficient update with variance correction \eqref{eq_new_varcor_coeff_update}. 

\begin{theorem}{(Restatement of \Cref{theo_loss_descent_red_var_cor})}\label{theo_loss_descent_full_red_cor_translated}
Under \Cref{assump:delta_bound}, if the local learning rate $0<\lambda\leq\frac{1}{12 Ls_*}$, then \Cref{alg:modified_FeDLRT_var_cor} leads to the global loss descent
\begin{align}
\begin{aligned}
          &\mathcal{L}(\Wr^{t+1}) - \mathcal{L}(\Wr^t)\leq
         - s_*\lambda (1-\delta^2 - 12 s_*\lambda L + \delta^2{s_*\lambda})   \norm{\augU^\top F(\augWr)\augV}^2 + L\vartheta,
\end{aligned}
     \end{align}
     with $\Wr^t=U^t S^t V^{t,\top}$ and $\Wr^{t+1}=U^{t+1} S^{t+1} V^{t+1,\top}$.
\end{theorem}
\begin{proof}

We split the adjusted coefficient update in \eqref{eq_new_varcor_coeff_update} into the non-augmented $r\times r$ matrix $S$  and the tree off-diagonal blocks given by the augmentation $\hatS$: 
\begin{align}\label{eq_hats_def}
    \hatS = \augS - \begin{bmatrix}
S & 0 \\
0 &0 
\end{bmatrix}.
\end{align}

Analogously to the proof of \Cref{theo_loss_descent_full_var_cor}, we consider 
\begin{align*}
    \mathcal{L}(\augWr^*)-\mathcal{L}(\augWr)&\leq \inner{\augWr^*-\augWr,F(\augWr)} + \frac{L}{2}\norm{\augWr^*-\augWr}^2\\
    &=  \inner{\augU\augS^*\augV^\top-\augU\augS\augV^\top,F(\augWr)} + \frac{L}{2}\norm{\augU\augS^*\augV^\top-\augU\augS\augV^\top}^2\\
    &=  \inner{\augS^* -\augS,\augU^\top F(\augWr)\augV} + \frac{L}{2}\norm{\augS^*-\augS}^2\\
     &=  \inner{\augS^* -\augS,-\nabla_\augS\mathcal{L}(\augWr)} + \frac{L}{2}\norm{\augS^*-\augS}^2,
\end{align*}
where the transformation uses orthonormality of $\augU$ and $\augV$ and definition of the projected gradient.
We split the right hand side in terms corresponding to 
augmented terms $\hatS$ and non-augmented terms $S$ according to~\eqref{eq_hats_def}, i.e.,
\begin{align}\label{helper_non_aug}
\inner{S^* -S,-\nabla_S\mathcal{L}(\augWr)} + \frac{L}{2}\norm{S^*-S}^2,
\end{align}
which is treated exactly as in the proof of \Cref{theo_loss_descent_full_var_cor}, and the augmented terms
\begin{align}\label{helper_aug}
\inner{\hatS^* -\hatS,-\nabla_\hatS\mathcal{L}(\augWr)} + \frac{L}{2}\norm{\hatS^*-\hatS}^2.
\end{align}

First we bound the term \eqref{helper_non_aug}.
Remember that $\hatS =0$ at the start of the local iterations due to orthonormality of $\augU,\augV$.
The coefficient update \eqref{eq_new_varcor_coeff_update} for $S$ reads
\begin{align}\label{helper_0001}
   S_c^{s+1} = S_c^s +  \lambda U^\top\left( F_c(\augWrc^s) - F_c(\augWr) + F(\augWr)\right)V.
\end{align}
Then we can readily apply \Cref{theo_loss_descent_full_var_cor} to obtain the bound 
\begin{align}\label{helper_non_aug_bound00}
   \inner{S^* -S,-\nabla_S\mathcal{L}(\augWr)} + \frac{L}{2}\norm{S^*-S}^2 \leq -( s_*\lambda)(1- 12( s_*\lambda)L) \norm{U^\top F(\augWr)V}^2.
\end{align}

Next, we bound \eqref{helper_aug}, starting with the first term:
\begin{align*}
    \inner{\hatS^* -\hatS,-\nabla_\hatS\mathcal{L}(\augWr)} & \overset{\text{(I)}}{=}\inner{\hatS^* -0,-\nabla_\hatS\mathcal{L}(\augWr)} \\
    &=
       \inner{ -\frac{\lambda}{C}\sum_{c=1}^C\sum_{s=0}^{s_*-1}\nabla_\hatS\mathcal{L}_c(\augWrc^s),-
       \nabla_\hatS\mathcal{L}(\augWr)} \\
   &=
   \frac{\lambda}{C}\sum_{c=1}^C\sum_{s=0}^{s_*-1}\inner{ \nabla_\hatS\mathcal{L}_c(\augWrc^s),
   \nabla_\hatS\mathcal{L}(\augWr)} \\
     &\leq
   \frac{\lambda}{C}\sum_{c=1}^C\sum_{s=0}^{s_*-1}\norm{ \nabla_\hatS\mathcal{L}_c(\augWrc^s)}\norm{
   \nabla_\hatS\mathcal{L}(\augWr)} \\
      &\overset{\text{(II)}}{\leq}
   \frac{\lambda}{C}\sum_{c=1}^C\sum_{s=0}^{s_*-1}\delta^2\norm{ \nabla_\augS\mathcal{L}(\augWr)}\norm{
   \nabla_\augS\mathcal{L}(\augWr)} \\
   & = \delta^2 s_* \lambda\norm{\nabla_\augS\mathcal{L}(\augWr)}^2 =  \delta^2 s_* \lambda\norm{\augU^\top F(\augWr)\augV}^2,
\end{align*}
where we use $\hatS=0$ in (I), and \Cref{assump:delta_bound} in (II).
Next, we bound the second term
\begin{align*}
    \frac{L}{2} \norm{\hatS^*-\hatS}^2 =&  \frac{L}{2}\norm{ -\frac{\lambda}{C}\sum_{c=1}^C\sum_{s=0}^{s_*-1}\nabla_\hatS\mathcal{L}(\augWrc^S)}^2\\
    \overset{\text{(I)}}{\leq} & \frac{L}{2} \lambda^2\frac{1}{C}\sum_{c=1}^C\norm{\sum_{s=0}^{s_*-1} \nabla_\hatS\mathcal{L}(\augWrc^S)}^2\\
     \overset{\text{(I)}}{\leq} & \frac{L}{2} {s_*}\lambda^2\frac{1}{C}\sum_{c=1}^C\sum_{s=0}^{s_*-1} \norm{\nabla_\hatS\mathcal{L}(\augWrc^S)}^2\\
    \leq & {s_*}\frac{L}{2} \delta^2\lambda^2\frac{1}{C}\sum_{c=1}^C\sum_{s=0}^{s_*-1}\norm{\nabla_\augS\mathcal{L}(\augWr)}^2 \\
    \leq & \frac{L}{2} \delta^2{(s_*\lambda)^2}\norm{\nabla_\augS\mathcal{L}(\augWr)}^2 =  \frac{L}{2}\delta^2{(s_*\lambda)^2}\norm{\augU^\top F(\augWr)\augV}^2 ,
\end{align*}
where we used Jensen's inequality in (I) again \Cref{assump:delta_bound}.
We combine the bound on the non-augmented terms \eqref{helper_non_aug_bound00} and the two bounds above for the augmented terms to
\begin{align*}
     &\mathcal{L}(\augWr^*)-\mathcal{L}(\augWr)\leq \inner{\augWr^*-\augWr,F(\augWr)} + \frac{L}{2}\norm{\augWr^*-\augWr}^2\\
     \leq & -( s_*\lambda)(1- 12( s_*\lambda)L) \norm{U^\top F(\augWr)V}^2 + \delta s_* \lambda \norm{\augU^\top F(\augWr)\augV}^2 +\delta{(s_*\lambda)^2} \norm{\augU^\top F(\augWr)\augV}^2 \\
    \overset{\text{(I)}}{\leq} & -( s_*\lambda)(1- 12( s_*\lambda)L) \norm{\augU^\top F(\augWr)\augV}^2 + \delta s_* \lambda \norm{\augU^\top F(\augWr)\augV}^2 +\delta{(s_*\lambda)^2} \norm{\augU^\top F(\augWr)\augV}^2 \\
      = & -( s_*\lambda)(1-\delta^2 - 12( s_*\lambda)L + \delta^2{(s_*\lambda)}) \norm{\augU^\top F(\augWr)\augV}^2,
\end{align*}
where we use in (I) $\norm{U^\top F(\augWr)V}\leq \norm{\augU^\top F(\augWr)\augV} $.
Using \Cref{eq_truncation_helperxxx}, we can conclude the proof:
\begin{align*}
          &\mathcal{L}(U^{t+1} S^{t+1} V^{t+1,\top}) - \mathcal{L}(U^t S^t V^{t,\top}) \\
          \leq & -( s_*\lambda)(1-\delta^2 - 12( s_*\lambda)L + \delta^2{(s_*\lambda)})  \norm{\augU^\top F(\augWr)\augV}^2 + L\vartheta.
     \end{align*}
\end{proof}

\subsection{Global convergence of FeDLRT with simplified variance correction}\label{sub_global_convergence_simple}
\begin{corollary}
{(Restatement of \Cref{theo_glob_convergence_simple})}
      Under \Cref{assump:delta_bound},  \Cref{alg:modified_FeDLRT_var_cor} guarantees for the learning rate $\lambda\leq\frac{1}{s_*(12 L +\delta^2)}$
    \begin{align}
          \min_{t=1,\dots,T}\norm{\nabla_\augS\mathcal{L}(\Wr^t)}^2\leq \frac{96 L}{T}\left(\mathcal{L}(\Wr^{1})-\mathcal{L}(\Wr^{{T+1}})\right) + {96 L^2}{\vartheta},
    \end{align}
     with $\Wr^t=U^t S^t V^{t,\top}$,  $\Wr^1=U^1 S^1 V^{1,\top}$. and $\Wr^{T+1}=U^{T+1} S^{T+1} V^{T+1,\top}$.
\end{corollary}
\begin{proof}

Consider \Cref{theo_loss_descent_red_var_cor}, 
\begin{align*}
 \mathcal{L}(\Wr^{t+1}) - \mathcal{L}(\Wr^{t})\leq 
        -( s_*\lambda)(1-\delta^2 - 12( s_*\lambda)L + \delta^2{(s_*\lambda)}) \norm{\augU^\top F(\augWr)\augV}^2 + L\vartheta
        \end{align*}
and assume that $\lambda s_* = \frac{1}{(12 L +\delta^2)}$, i.e. $\lambda=\frac{1}{s_*(12 L +\delta^2)}\leq\frac{1}{Ls_*}$, which obeys the learning rate requirement of Theorem~\ref{theo_loss_descent_full_var_cor}. Plugging this learning rate into \eqref{eq_helper123097213} gives
\begin{align*}
          \norm{\nabla_\augS\mathcal{L}(\Wr^{t})}^2\leq 96 L \left(\mathcal{L}(\Wr^t)-\mathcal{L}(\Wr^{t+1})+ L\vartheta\right),
\end{align*}
where we use $(\frac{1}{4}-\delta^2)\leq \frac{1}{4}$ and $\frac{1}{(12 L +\delta^2)}\leq \frac{1}{12 L}$
Averaging from $t=1$ to $t=T$ yields
\begin{align*}
     \min_{t=1,\dots,T}\norm{\nabla_\augS\mathcal{L}(\Wr^{t})}^2
     \leq&\frac{1}{T}\sum_{t=1}^T\norm{\augU^\top F(\augWr)\augV}^2\\
     \leq& \frac{96 L}{T}\left(\mathcal{L}(\Wr^{t=1})-\mathcal{L}(\Wr^{t={T+1}})\right) + {96 L^2}{\vartheta},
\end{align*}
which concludes the proof.
\end{proof}

\end{document}